\documentclass{article}

\usepackage{microtype}
\usepackage{graphicx}
\usepackage{subfigure}
\usepackage{booktabs} 
\usepackage{multirow}
\usepackage{tabularx}
\usepackage{longtable}
\usepackage{hyperref}

\usepackage[accepted]{icml2024}

\usepackage{amsmath}
\usepackage{amssymb}
\usepackage{mathtools}
\usepackage{amsthm}
\usepackage{bm}
\usepackage{enumitem}
\usepackage{rotating}

\setitemize[1]{itemsep=-2pt,partopsep=0pt,parsep=\parskip,topsep=2pt}
\usepackage[capitalize,noabbrev]{cleveref}

\theoremstyle{plain}
\newtheorem{theorem}{Theorem}[section]

\theoremstyle{definition}

\theoremstyle{remark}

\usepackage[textsize=tiny]{todonotes}

\icmltitlerunning{Parameter Efficient Quasi-Orthogonal Fine-Tuning via Givens Rotation}

\begin{document}

\twocolumn[
\icmltitle{Parameter Efficient Quasi-Orthogonal Fine-Tuning via Givens Rotation}



\icmlsetsymbol{equal}{*}

\begin{icmlauthorlist}
\icmlauthor{Xinyu Ma}{pku,lab}
\icmlauthor{Xu Chu}{pku,lab,cfcs}
\icmlauthor{Zhibang Yang}{lab}
\icmlauthor{Yang Lin}{pku,lab}
\icmlauthor{Xin Gao}{pku,lab}
\icmlauthor{Junfeng Zhao}{pku,lab}
\end{icmlauthorlist}

\icmlaffiliation{pku}{School of Computer Science, Peking University, Beijing, China}
\icmlaffiliation{lab}{Key Laboratory of High Confidence Software Technologies, Ministry of Education, Beijing, China}
\icmlaffiliation{cfcs}{Center on Frontiers of Computing Studies, Peking University, Beijing, China}
\icmlcorrespondingauthor{Xu Chu}{chu\_xu@pku.edu.cn}

\icmlkeywords{Machine Learning, ICML}

\vskip 0.3in
]



\printAffiliationsAndNotice{} 

\begin{abstract}
With the increasingly powerful performances and enormous scales of pretrained models, promoting parameter efficiency in fine-tuning has become a crucial need for effective and efficient adaptation to various downstream tasks.
One representative line of fine-tuning methods is Orthogonal Fine-tuning (OFT), which rigorously preserves the angular distances within the parameter space to preserve the pretrained knowledge.
Despite the empirical effectiveness, OFT still suffers low parameter efficiency at $\mathcal{O}(d^2)$ and limited capability of downstream adaptation.
Inspired by Givens rotation, in this paper, we proposed quasi-Givens Orthogonal Fine-Tuning (qGOFT) to address the problems. 
We first use $\mathcal{O}(d)$ Givens rotations to accomplish arbitrary orthogonal transformation in $SO(d)$ with provable equivalence, reducing parameter complexity from $\mathcal{O}(d^2)$ to $\mathcal{O}(d)$. Then we introduce flexible norm and relative angular adjustments under soft orthogonality regularization to enhance the adaptation capability of downstream semantic deviations.
Extensive experiments on various tasks and pretrained models validate the effectiveness of our methods.
\end{abstract}

\section{Introduction} \label{sec:intro}
Pretrained foundation models like GPT-3 \cite{brown2020language}, LLaMA-2 \cite{touvron2023llama} and ViT \cite{dosovitskiy2020image} have exhibited remarkable performances across various NLP and vision tasks \cite{devlin2018bert, liu2019roberta, zhai2019large}, which can be attributed to the dramatic increase in model scales \cite{wei2022emergent}. 
However, the explosive parameter growth poses an increasingly huge challenge when adapting these models to diverse downstream tasks or expertise domains. Thus, parameter-efficient fine-tuning (PEFT) mechanisms have become an urgent need for downstream adaptation and motivate a long line of research \cite{lester2021power, li2021prefix, houlsby2019parameter, hu2021lora, mao2022unipelt}.
Amongst them, \textit{reparameterization tuning} \cite{hu2021lora, edalati2022krona, zhang2022adaptive} has been most widely adopted, where the pretrained model remains frozen and only the updates of partial modules are reparameterized with low parameter budgets and fine-tuned. 

The practical principle of reparameterization tuning assumes that the tuned model parameters should not drift too far from the original model (e.g., zero initializations and small learning rates of LoRA-series methods \cite{hu2021lora, zhang2022adaptive, dettmers2023qlora}).
We summarize the core idea behind this manner is that most pretrained concepts and measurements in the latent metric space should be preserved after fine-tuning. 
This motivating insight thereby inspires some research to consider an isometric mapping on the original parameters while preserving certain underlying distance metrics to conduct adaptation. 
Among various metrics, angular distance metrics (e.g., cosine similarity) are commonly applied in neural latent spaces \cite{wang2018cosface, jin2022domain, li2023distributional, xu2023seqcare, ma2024fused}, due to its outstanding capability of encoding semantic information \cite{liu2018learning, liu2021orthogonal, ma2023patient}. This motivates a line of popular fine-tuning methods, dubbed Orthogonal Fine-tuning (OFT) \cite{qiu2023controlling}, which conducts angle-preserving mapping on the original linear weights using orthogonal transformation.
As the weight vectors in pretrained models depict the way of projecting latent semantics, this angular preserving property ensures OFT with invariant relative semantic distances between concepts after the fine-tuned projection. With this merit, OFT has demonstrated promising performances for fine-tuning text-to-image diffusion models \cite{qiu2023controlling}.

Despite the empirical success of OFT, there remain two crucial problems to be solved. The first problem is that OFT methods are essentially NOT parameter-efficient. Due to the high dimensionality of linear layers in pretrained models, the number of trainable parameters (i.e., the size of the orthogonal matrix) in OFT can be as excessively large as $\mathcal{O}(d^2)$, where $d$ is the hidden dimension of the linear layer.
To address the problem, \citet{qiu2023controlling} assumed a block diagonal structure of the orthogonal matrix to reduce the parameter cost to $\mathcal{O}(d^2/N)$, where $N$ denotes the number of blocks. However, the saving of parameters comes with the price of sacrificing expressiveness that it can only represent a limited subset of orthogonal transformations in $\mathbb{R}^{d}$. To this end, a natural question is: \textbf{Q1. \textit{Is there a parameter-efficient approach to depict a learnable orthogonal matrix without losing expressiveness?}}

The second problem is that OFT has a limited capability of adapting semantic shifts from downstream tasks. Though the majority of pretrained semantic knowledge is encouraged to be preserved when performing downstream adaptations, there still exists some \textbf{slight} semantic shifts between pretrained and downstream corpus, depicted by the deviations in relative semantic distances between concepts and attention distribution shifts. However, OFT strictly ensures the relative angular distances between weight vectors and their norms are unchanged, which are strongly related to the semantic distances and attention distributions. These factors make OFT incompatible with learning those subtle semantic shifts.
Therefore, we are curious that: \textbf{Q2. \textit{How to increase the adaptation capability of OFT without immensely increasing the parameter complexity?}} 

\textbf{Addressing Q1:} In this work, we try to comprehend OFT from a geometric view instead of algebraic conventions \cite{cayley1846quelques, qiu2023controlling}. Geometrically, orthogonal transformation is either rotating the linear space (i.e., special orthogonal group $SO(d)$ with determinant $1$) or conducting reflection after rotations (i.e., roto-reflection mappings with determinant $-1$). 
However, roto-reflection mappings may hugely switch the semantics of neurons in the latent space (e.g., the further from the reflection plane, the greater the semantic change). This contradicts the practical principle of PEFT methods, which aims to prevent the tuned model from drifting too far from the base model, potentially resulting in significant performance degradation.
Hence, we mainly focus on the rotational OFT in $SO(d)$. 

Our insight comes from Givens rotation \citep[Section 11.3.1]{press2007numerical}, which rotates the plane subspace spanned by two of the entire coordinate axes. Givens rotation is naturally an orthogonal matrix with one degree of freedom - the rotating angle and is adopted as an effective tool for approximating orthogonal matrix \cite{frerix2019approximating}, matrix QR decomposition \cite{rath1982fast} and linear least squares problem \cite{george1980solution}, etc. This inspires us with a motivating idea: 
\textit{Is it possible to apply a combination of several Givens rotations to express any orthogonal transformation in $SO(d)$? }
In this work, we have proved that in a $d$-dimensional linear space, it requires at most $d-1$ specific Givens rotations to rotate a vector $\bm{x} \in \mathbb{R}^d$ to any vector $\bm{y} \in \mathbb{R}^d$ on the same sphere with $\bm{x}$. 
Based on this theoretical result, we thereby propose a novel Givens OFT (GOFT) method with high parameter efficiency. 
GOFT only requires the product of $d-1$ Givens rotations to express any rotation, each of which requires one learnable parameter, namely the rotation angle. Thus, we use only \underline{$\bm{\mathcal{O}(d)}$} learnable parameters to efficiently achieve the equivalent expressive power of OFT in $SO(d)$.
However, sequentially multiplying $d-1$ Givens rotations can increase the computational costs intensely. To mitigate this problem, we have altered the rotation strategy by introducing a novel parallel rotation strategy that applies only \underline{$\bm{\mathcal{O}(\log d)}$} sparse matrix multiplications while maintaining the same expressiveness.

\textbf{Addressing Q2:} 
We aim to incorporate slightly tunable relative angles and flexible norms for better adaptation capability.
With Givens rotation, we decompose the orthogonal transformation into $\mathcal{O}(d)$ individual 2-dimensional rotations, making it possible to incorporate adjustable flexibility to every single Givens rotation. Therefore, we further propose a quasi-Givens OFT method dubbed qGOFT, which introduces $\mathcal{O}(d)$ of extra trainable parameters and relaxes the strict Givens rotation with adjustable norms and tunable relative angles with an orthogonality regularizer. 
With this design, qGOFT allows for a better adaptation to the slight semantic shift of downstream tasks and domains while preserving the angular measurements as much as possible.

To summarize, we highlight our contributions as follows. Inspired by Givens rotation, we proposed a parameter-efficient quasi-orthogonal fine-tuning method dubbed qGOFT, which mainly enhances OFT from two aspects:
\begin{itemize}[leftmargin=*]
    \vspace{-0.05in}
    \item \textbf{Enhancing Parameter Efficiency with Equivalent Expressiveness}: We designed a Givens-based OFT (GOFT) method that reduces the parameter complexity from quadratic ($\mathcal{O}(d^2)$) to linear ($\mathcal{O}(d)$) with expressive power equivalent to OFT in $SO(d)$. To further achieve a better trade-off between parameter and computational costs, we introduce a novel parallel rotation strategy to increase computation efficiency, reducing the number of sparse matrix multiplications from $\mathcal{O}(d)$ to $\mathcal{O}(\log d)$.
    \item \textbf{Enhancing Adaptation Capability}: Based on GOFT, we further introduced quasi-Givens rotations and proposed the qGOFT method. qGOFT enables adjustable vector norms and slightly tunable angular measurements, which improves the adaptation capability to the semantic shift underlying downstream tasks and various domains.
\end{itemize}
\vspace{-0.05in}
Extensive experiments are conducted on a wide range of NLP and vision tasks with various pretrained models.
Results demonstrate the effectiveness of our methods with outstanding performances under low parameter budgets.

\section{Related Works}
As pretrained foundation models have become increasingly powerful and heavily parameterized, it draws huge attention from researchers to developing more parameter-efficient approaches to fine-tuning them for downstream adaptations. Rather than requiring an entire fine-tuned parameter set \cite{howard2018universal}, parameter-efficient fine-tuning (PEFT) develops a lightweight adapter for different downstream tasks, allowing for a dramatically lower model training and storage cost \cite{lester2021power}.
There are three mainstream approaches of PEFT \cite{lialin2023scaling}: \textit{prompt tuning} \cite{lester2021power, li2021prefix, liu2022p,liu2021pv2}, where additional learnable prefix tokens are concatenated to the input of Transformer layers; \textit{adapter tuning} \cite{houlsby2019parameter, pfeiffer2020adapterfusion, he2022sparseadapter, liu2022few}, where additional trainable modules are inserted to the original model; and \textit{reparameterization tuning} \cite{aghajanyan2021intrinsic, hu2021lora, zhang2022adaptive, dettmers2023qlora, edalati2022krona, lin2024lora}, where the model architecture remains unchanged and only the delta of partial model parameters are reparameterized with low parameter budgets and get fine-tuned.
There are also some works \cite{he2021towards, mao2022unipelt, chen2022parameter} trying to integrate the three paradigms with one unified framework.
Among those PEFT methods, reparameterization tuning is most relevant to our work. LoRA-based methods \cite{hu2021lora, zhang2022adaptive, dettmers2023qlora} apply slight additive low-rank updates to linear weights in pretrained models. Based on the metric preservation insight from reparameterization tuning, isometry-based methods, typically orthogonal fine-tuning (OFT) \cite{qiu2023controlling}, have been further proposed. OFT preserves the pretrained semantics and concepts via conducting angular-preserving mapping to the linear weights.
However, OFT suffers the obstacles of heavy parameter costs and limited capability of adapting downstream semantic deviations. 

In this work, we further proposed qGOFT to overcome those challenges, making OFT parameter-efficient by reducing parameter costs from quadratic to linear and meanwhile, allowing for more flexible adjustment on relative semantic deviations under soft orthogonality constraints.

\vspace{-0.05in}
\section{Preliminaries}
\subsection{Orthogonal Fine-Tuning} \label{sec:OFT_pre}
\begin{figure}[h]
    \centering
    \vspace{-0.05in}
    \includegraphics[width=\linewidth]{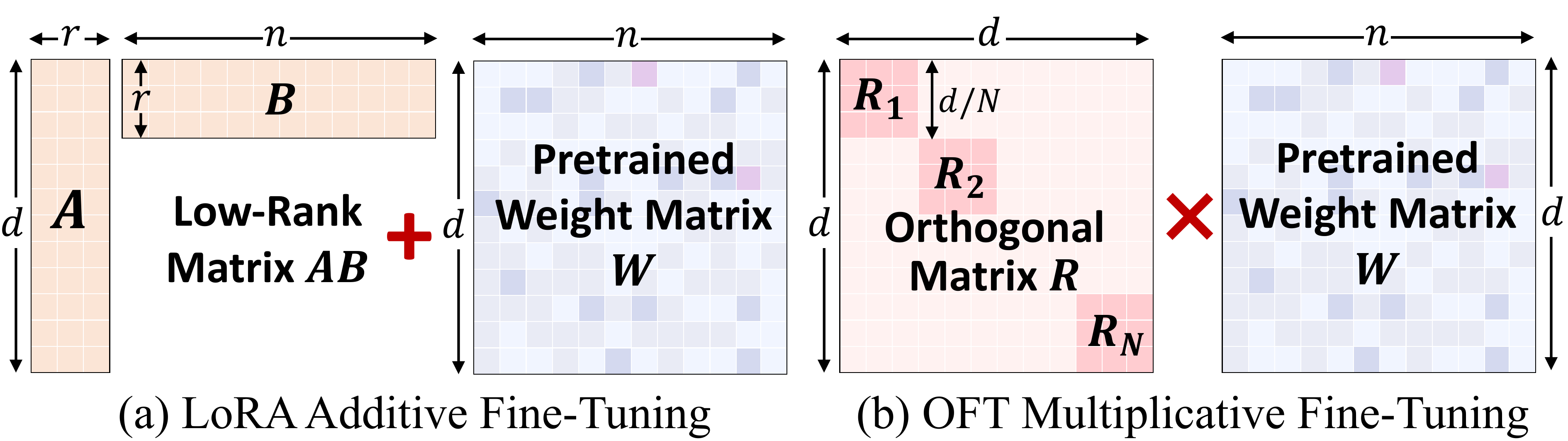}
    \vspace{-0.2in}
    \caption{LoRA and OFT Reparameterization Tuning Methods.}
    \label{fig:oft_pre}
\end{figure}

The original OFT \cite{qiu2023controlling} is basically a reparameterization tuning method like LoRA. The difference is that OFT applies multiplicative updates to the frozen pretrained weight matrices, whereas LoRA conducts additive updates with low-rank delta weights (see Fig.~\ref{fig:oft_pre}). 
OFT requires the multiplier to be strictly orthogonal, and the motivation comes from preserving the angular distance metrics in the latent space, such that the pretrained knowledge can be highly preserved.
In detail, OFT reparameterizes the new linear weight matrix $\bm{W}_*$ as the product of a learnable orthogonal matrix $\bm{R}$ and the original frozen weight matrix $\bm{W}$, and the forward pass is modified from $\bm{h} =  \bm{W}^\top \bm{x}$ to $\bm{h} = \bm{W_*}^\top \bm{x} = (\bm{RW})^\top \bm{x}$, where $\bm{x} \in \mathbb{R}^{d}$ and $\bm{h} \in \mathbb{R}^{n}$ are the input and output vectors, respectively.
The orthogonality of $\bm{R}$ is ensured by Cayley parameterization trick \cite{cayley1846quelques}, i.e., $\bm{R} = (\bm{I} + \bm{Q})(\bm{I} - \bm{Q})^{-1}$, where $\bm{Q}$ is a skew-symmetric matrix (i.e., $\bm{Q} + \bm{Q}^\top = \bm{0}$). 
By this means, the parameters to be tuned in a single layer are of the size of matrix $\bm{Q}$, which is $\mathcal{O}(d^2)$.
For better parameter efficiency, \citet{qiu2023controlling} sacrifice some expressiveness and introduce a block-diagonal structure of $\bm{R}$, i.e., $\bm{R} = \mathrm{diag}(\bm{R_1}, ..., \bm{R_N})$, where $N$ is the number of diagonal orthogonal blocks also parameterized with the Cayley trick. 

However, as we mentioned in the Introduction, the original OFT suffers from two crucial problems: 1) the inevitable trade-off between parameter efficiency and expressive power, and 2) the limited capability of adapting to downstream semantic shifts. In the next section, we will introduce our detailed method to address both two problems.

\subsection{Givens Rotation} \label{sec:givens_pre}
In numerical linear algebra, a Givens rotation \citep[Section 11.3.1]{press2007numerical} is a rotation in the plane spanned by two coordinate axes. Algebraically, a Givens rotation is represented by an orthogonal matrix of the form:
\begin{equation}
    \bm{G}(i,j;\theta) = \begin{bmatrix}
    1 & \cdots & 0 & \cdots & 0 & \cdots & 0 \\ 
    \vdots & \ddots & \vdots &  & \vdots &  & \vdots \\ 
    0 & \cdots & \cos{\theta} & \cdots & -\sin{\theta} & \cdots & 0 \\ 
    \vdots & & \vdots & \ddots & \vdots &  & \vdots \\
    0 & \cdots & \sin{\theta} & \cdots & \cos{\theta} & \cdots & 0 \\ 
    \vdots & & \vdots &  & \vdots & \ddots & \vdots \\
    0 & \cdots & 0 & \cdots & 0 & \cdots & 1 \\ 
\end{bmatrix},
\end{equation}
where $\cos{\theta}$ and $\sin{\theta}$ appear at the intersection positions of $i$-th and $j$-th rows and columns, and the other non-zero entries are at the diagonal with all $1$s. In other words, the non-zero elements $g_{mn}$ in $\bm{G}(i,j;\theta)$ are given by: 



\vspace{-0.15in}
\begin{equation}
\left\{ \begin{array}{*{35}{l}}
    & g_{kk} = 1, & \quad \mathrm{for} \ k \neq i,j; \\
    & g_{kk} = \cos{\theta}, & \quad \mathrm{for} \ k = i,j; \\
    & g_{ij}  = -g_{ji} = -\sin{\theta}.    \\
\end{array} \right.
\vspace{-0.05in}
\end{equation}
Geometrically, for a vector $\bm{x} \in \mathbb{R}^d$ in an $d$-dimensional linear space, the product $\bm{G}(i,j;\theta)\bm{x}$ represents a counterclockwise rotation of the vector $\bm{x}$ in the subspace plane spanned by $i$-th and $j$-th coordinate axes with angle  $\theta$.

\section{Methodology}
In this section, we formally introduce the detailed theoretical results and methods to address the problems of OFT as we mentioned in the Introduction (i.e., \textbf{Q1} and \textbf{Q2} in Sec.\ref{sec:intro}).

\subsection{Q1: Enhancing Parameter Efficiency of OFT in $SO(d)$ with Equivalent Expressive Power} \label{sec:method_1}

\begin{figure*}
    \centering
    \includegraphics[width=0.9\linewidth]{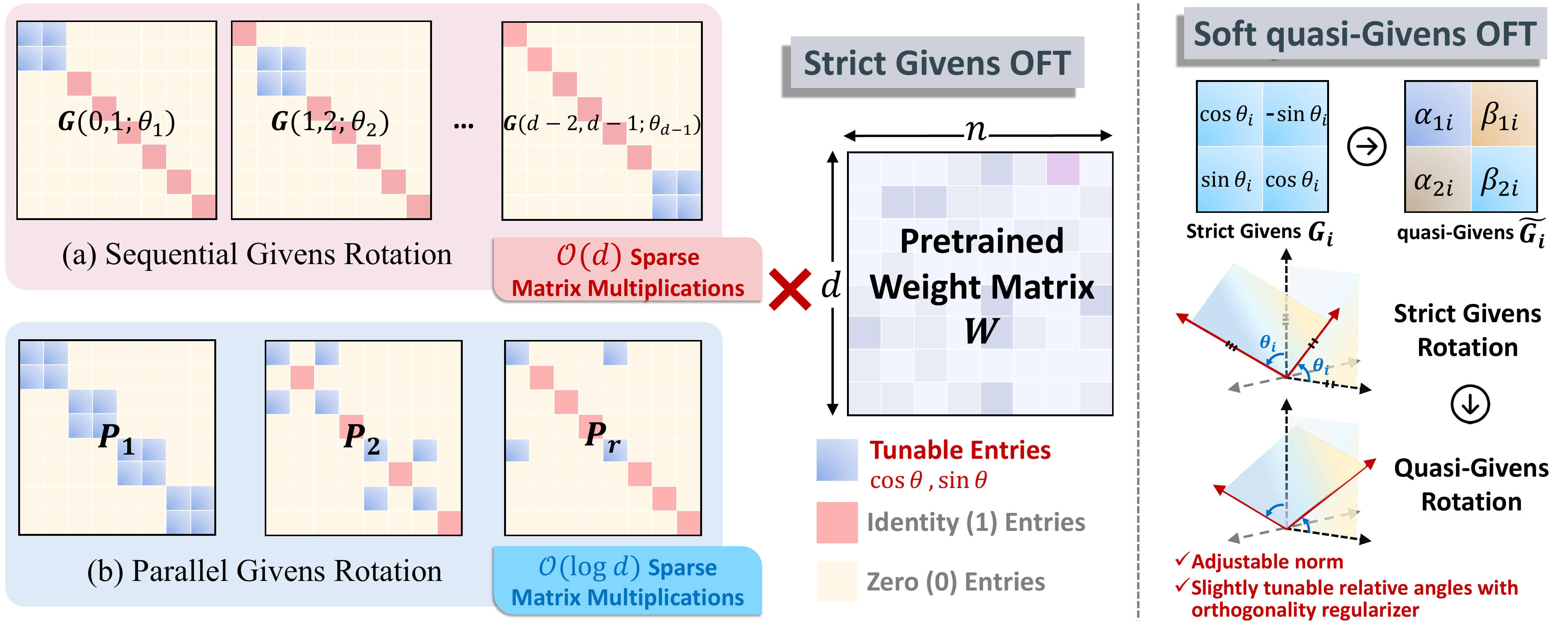}
    \vspace{-0.1in}
    \caption{Our proposed method: quasi-Givens Orthogonal Fine-Tuning (qGOFT). The left subfigure denotes the strict GOFT which applies $d-1$ Givens rotation to left-multiply with the pretrained weight matrix, where (a) and (b) depict the sequential and parallel rotation manner, respectively. The right subfigure illustrates how qGOFT works, where each Givens rotation in GOFT is substituted with a quasi-Givens matrix for norm and angular relaxation.}
    \label{fig:method}
    \vspace{-0.1in}
\end{figure*}

As we introduced in Section \ref{sec:OFT_pre}, the original OFT applies an orthogonal transformation to the frozen pretrained linear weight matrix to conduct downstream adaptations, where the orthogonality is ensured by the Cayley parameterization trick at the cost of $\mathcal{O}(d^2)$ parameters. In this subsection, we are attempting to reduce the parameter complexity to accomplish an equivalent orthogonal transformation in $SO(d)$ without losing any expressiveness.

Let's start with considering the geometric meaning of $SO(d)$ orthogonal transformations. The basic idea of $SO(d)$ is to rotate the linear space. Hence, in OFT, our focus is specifically on determining the rotation angles. This prompts the question: is it truly necessary to have $\mathcal{O}(d^2)$ degrees of freedom to represent rotation angles in a $d$-dimensional linear space?
From this point of view, we draw inspiration from the Givens rotation introduced in Section \ref{sec:givens_pre}, which is in essence rotating the plane spanned by two of the entire $d$ dimensions while keeping the other axes unchanged. Given two fixed axes, we can easily find there is only one degree of freedom for a Givens rotation, namely the rotation angle. Therefore, we are wondering if it is possible to fit any rotation in the $d$-dimensional space with a combination of less than $\mathcal{O}(d^2)$ Givens rotations, thereby reducing the degrees of freedom to express an $SO(d)$ orthogonal transformation. Guaranteed by the following theoretical result, we argue that we only need at most $d-1$ degree of freedom to accomplish any $d$-dimensional rotation.

\begin{theorem}
    Given any vector $\bm{x} \in \mathbb{R}^d$, there always exist $d-1$ Givens rotations $\{\bm{G}(i_k, j_k; \theta_k)\}_{k=1}^{d-1}$ that can transform $\bm{x}$ to any vector $\bm{y} \in \mathbb{R}^d$ on the same sphere with $\bm{x}$, i.e., $\prod_{k=1}^{d-1} \bm{G}(i_k, j_k; \theta_k) \bm{x} = \bm{y}$, satisfying $||\bm{y}||_2^2 = ||\bm{x}||_2^2$.
    \label{theo:givens}
\end{theorem}
\vspace{-0.2in}
\begin{proof}
    Without losing generality, let us assume $||\bm{x}||_2 = 1$ and $\bm{y} = [1, 0, 0, \cdots, 0]$ and denote $\bm{G}_k := \bm{G}(i_k, j_k; \theta_k)$. We below construct $d-1$ Givens rotations satisfying $\prod_{k=1}^{d-1} \bm{G}_k \bm{x} = \bm{y}$ (See illustrative example in Fig.\ref{fig:proof}, Appendix \ref{apdx:figs}). The first Givens rotation aims to rotate the plane spanned by the last two axes (i.e., $(d-2)$-th and $(d-1)$-th axes) such that the last dimension of $\bm{x}$ is rotated to 0, i.e., $\bm{x}[d-1] = 0$. Hence the first Givens rotation $\bm{G}_1$ can be represented by: $i_1 = d-2, j_1 = d-1, \theta_1 = -\arccos{\frac{\bm{x}[d-2]}{\sqrt{\bm{x}[d-2]^2+\bm{x}[d-1]^2}}}$. Then the second Givens rotation aims to rotate the $(d-3, d-2)$ plane such that the $(d-2)$-th dimension of $\bm{x}$ is rotated to 0, and so forth. We denote the vector after the $k$-th rotation $\bm{x}_k$, then the following Givens rotation follows the same paradigm that $\bm{G}_k$ rotates $(d-k-1, d-k)$ plane such that $\bm{x}_k[d-k] = 0$, which is $i_k = d-k-1, j_k = d-k, \theta_k = -\arccos{\frac{\bm{x}_k[d-k-1]}{\sqrt{\bm{x}_k[d-k-1]^2+\bm{x}_k[d-k]^2}}}$.  Finally, after $d-1$ times of such Givens rotation, all the dimensions of $\bm{x}$ except the first one have been rotated to $0$. Moreover, all the Givens rotations are orthogonal matrices where the norm of the vector will not be modified after the rotation. Therefore, following this design, we have $\prod_{k=d-1}^{0} \bm{G}_k \bm{x} = [1, 0, 0, \cdots, 0] = \bm{y}$.
\end{proof}

This theorem gives an upper bound of $d-1$ degrees of freedom when depicting a rotation in a $d$-dimensional space. More importantly, the proof of this theorem essentially explains the way to express any rotational orthogonal transformation with only $d-1$ parameters. 
Therefore, based on the proof, we designed a novel parameter-efficient Givens OFT method as illustrated in Fig.\ref{fig:method}(a). 
Specifically, we applied $d-1$ Givens rotation matrices $\{\bm{G}_k\}_{k=1}^{d-1}$, where each matrix $\bm{G}_k$ rotates the plane spanned by $(k-1)$-th and $k$-th axes with only one learnable parameter - the rotation angle. In other words, we have $\bm{G}_k = \bm{G}(k-1, k; \theta_k)$, where $\theta_k$ is a learnable parameter. 
Sequentially calculating the products of $\prod_{k=1}^{d-1} \bm{G}_k$, we can obtain an orthogonal matrix parameterized by $d-1$ angular parameters $\{\theta_k\}_{k=1}^{d-1}$, whose expressiveness is equivalent to any rotational transformation according to Theorem \ref{theo:givens}.
Therefore, we have proposed a novel OFT method which is 1) \textbf{parameter-efficient}, reducing the parameter complexity from quadratic $\mathcal{O}(d^2)$ to linear $\mathcal{O}(d)$; and 2) \textbf{without losing expressiveness}, as the products of $d-1$ Givens rotations $\bm{G}(k-1,k;\theta_k)$ can express any orthogonal transformations in $SO(d)$.

\textbf{Improving Computational Efficiency:} However, sequentially conducting $\mathcal{O}(d)$ Givens rotations also takes $\mathcal{O}(d)$ times of matrix multiplications, and this intensively increases the computational costs. 
To balance the trade-off between parameter efficiency and computational costs, as demonstrated in Fig.\ref{fig:method}(b), we proposed a novel GPU-friendly parallel rotation strategy without introducing extra learnable parameters, which reduces the required number of multiplications from $\mathcal{O}(d)$ to $\mathcal{O}(\log d)$. 
Specifically, similar to proving Theorem \ref{theo:givens}, we first conduct $d/2$ non-overlapped Givens rotations in parallel (i.e., $\bm{P}_1$ in Fig.\ref{fig:method}(b)), which simultaneously rotate all the planes spanned by $(2k)$-th and $(2k+1)$-th axes to ensure all $(2k+1)$-th axes are rotated to the desired positions. Then the second transformation (i.e., $\bm{P}_2$) comprises $d/4$ simultaneous non-overlapped Givens rotations, rotating all the planes spanned by $(4k)$-th and $(4k+2)$-th axes to ensure all $(4k+2)$-th axes are rotated to the desired positions, and so forth (See Fig.\ref{fig:parallel}, Appendix \ref{apdx:figs} for illustrative examples). By this means, we can also rotate a vector to any position on the same sphere with $d-1$ parameter costs in total.
Formally, denoting each transformation as $\bm{P}_r, r= 1,2,\cdots, \log d$, we have:
\begin{equation}
    \bm{P}_r = \prod_{k=0}^{(d/2^r)-1} \bm{G}(2^r k, 2^{r-1}(2k+1); \theta_k^r).
    \label{eq:parallel}
\end{equation}
It is worth mentioning that the products in $\bm{P}_r$ do not need real matrix multiplication operations, as those rotations are \textbf{non-overlapped}. Thus, they can be easily implemented by directly setting the values on the corresponding coordinates as shown in Fig.\ref{fig:method}(b).
Therefore, by calculating $\mathcal{O}(\log d)$ sparse matrix multiplications of $\bm{P}_r$, we can obtain an arbitrary orthogonal transformation on the latent linear space. We name this method GOFT (Givens Orthogonal Fine-Tuning), with the forward pass:
\begin{equation}
    \bm{h} = (\bm{RW})^\top \bm{x} = \left(\left(\prod_{r=1}^{\log d} \bm{P}_r \right) \bm{W} \right)^\top \bm{x}.
    \label{eq:fw_pass}
\end{equation}
In a word, GOFT achieves the same expressiveness as rotational OFT while reducing the parameter efficiency from $\mathcal{O}(d^2)$ to $\mathcal{O}(d)$. With the design of parallel rotations, GOFT further improves the computational efficiency with only $\mathcal{O}(\log d)$ sparse matrix multiplications.

\subsection{Q2: Enhancing Adaptation Capability}
The original OFT ensures the norms of vector neurons and the relative angles between them remain strictly preserved. This can limit the adaptation capability of OFT, as the norms and relative angular measurements are important factors in representing the relative semantic and conceptual shifts across various domains. Therefore, we try to incorporate those two factors to relax the strict orthogonality in OFT for better downstream adaptation.

However, as Cayley parameterization \cite{cayley1846quelques} (i.e., $\bm{R} = (\bm{I} + \bm{Q})(\bm{I} - \bm{Q})^{-1}$, $\bm{Q}$ is skew-symmetric) enforces the matrix to be strictly orthogonal, the relaxation cannot be accomplished intuitively in OFT. While based on GOFT, we decompose the orthogonal transformation to $\mathcal{O}(d)$ rotations which allows for more flexible adjustments on each rotation. Hence, in this subsection, we introduced quasi-Givens OFT (qGOFT), which relaxes every Givens rotation with adjustable norms and flexible relative angular tuning under soft orthogonality constraints.

Specifically, as illustrated in the right subfigure in Fig.\ref{fig:method}, we altered every Givens rotation $\bm{G}_i$ to a quasi-Givens transformation $\tilde{\bm{G}_i}$. Taking the submatrix spanned by the two rotated coordinate axes, we have:
\begin{equation}
    \bm{G}_i = \begin{bmatrix}
        \cos \theta_i & -\sin \theta_i \\
        \sin \theta_i & \cos \theta_i
    \end{bmatrix} \rightarrow
    \tilde{\bm{G}_i} = \begin{bmatrix}
        \alpha_{1i} & \beta_{1i} \\
        \alpha_{2i} & \beta_{2i}
    \end{bmatrix} = (\bm{\alpha}_i, \bm{\beta}_i).
\end{equation}
We can see that $\tilde{\bm{G}_i}$ is currently an arbitrary 2-dimensional linear mapping with four learnable parameters (i.e., $\bm{\alpha}$ and $\bm{\beta}$). However, we do not want to immensely adjust the angular metric space of the original linear weights, as there is already abundant pretrained knowledge (e.g., relative semantic distances between words) underlying the original angular metric space, and all we need is a slight fine-tune. Hence, we propose tuning $\tilde{\bm{G}_i}$ with a soft orthogonality constraint, which encourages the two column vectors of $\tilde{\bm{G}_i}$ (i.e., $\bm{\alpha}_i, \bm{\beta}_i$) to be as orthogonal as possible.  
Practically, we apply a regularizer to draw $\langle \bm{\alpha}_i, \bm{\beta}_i \rangle$ close to $0$, which penalizes $\sum_i^{d-1} \langle \bm{\alpha}_i, \bm{\beta}_i \rangle ^2$ in the final training objective with strength $\lambda$, a hyper-parameter to be tuned. 

Moreover, considering the norm adjustments, given a vector $\bm{x}$ with transformation $\tilde{\bm{G}_i}$, the squared norm of the transformed vector $\tilde{\bm{G}_i} \bm{x}$ is:
\begin{align}
    \bm{x}^\top \tilde{\bm{G}_i}^\top \tilde{\bm{G}_i} \bm{x} & = \bm{x}^\top \begin{bmatrix}
    \alpha_{1i}^2+\alpha_{2i}^2 & \alpha_{1i}\beta_{1i}+\alpha_{2i}\beta_{2i} \\
    \alpha_{1i}\beta_{1i}+\alpha_{2i}\beta_{2i} & \beta_{1i}^2+\beta_{2i}^2
    \end{bmatrix} \bm{x} \nonumber\\
     & =  \bm{x}^\top  \begin{bmatrix}
    ||\bm{\alpha}_i||_2^2 & \langle \bm{\alpha}_i, \bm{\beta}_i \rangle \\
    \langle \bm{\alpha}_i, \bm{\beta}_i \rangle & ||\bm{\beta}_i||_2^2
    \end{bmatrix}  \bm{x}.
\end{align}
We can see that under orthogonality constraints of $(\bm{\alpha}_i, \bm{\beta}_i)$, the norm adjustment has been accomplished with the diagonal entries of $\tilde{\bm{G}_i}^\top \tilde{\bm{G}_i}$ and we do not conduct further restrictions on it. Though qGOFT introduces 3 times more learnable parameters than GOFT, it is still of an $\mathcal{O}(d)$ complexity, far more efficient than $\mathcal{O}(d^2)$ of the original OFT.

\begin{table*}[t]
\vspace{-0.1in}
\caption{Experimental results with DeBERTaV3-base on GLUE development set. The best results on each dataset are shown in \textbf{bold}. We report the mismatched accuracy for MNLI, Matthew's correlation for CoLA, averaged correlation for STS-B, and accuracies for other tasks. We report the mean results of 5 runs with different random seeds.}
\begin{center}
\begin{small}
\begin{tabular}{l|c|ccccccccc}
\toprule
\multirow{2}{*}{\textbf{Method}}&\multirow{2}{*}{\textbf{\#Param.}} &\textbf{MNLI} &\textbf{ SST-2}&\textbf{CoLA} &\textbf{QQP}&\textbf{QNLI}&\textbf{RTE}&\textbf{MRPC}&\textbf{STS-B}& \textbf{All} \\
 && mm-acc & Acc & Mcc& Acc &Acc &Acc &Acc &Corr &Ave.\\
\midrule
\textup{Full Finetuning} &184M   & 90.12 & 95.63& 69.19 & \textbf{92.40} & 94.03& 83.75 & 89.46 & 91.60& 88.27 \\
\textup{BitFit} &0.1M   & 89.91 & 94.84& 66.96 & 88.41 & 92.24& 78.70 & 87.75 & 91.35& 86.27 \\
\textup{H-Adapter} &0.31M   & 90.02 & 95.41& 67.65 & 91.54 & 93.52& 83.39 & 89.25 & 91.31& 87.76 \\
\textup{P-Adapter} &0.30M   & 90.06 & 94.72 & 69.06 & 91.40 & 93.87 & 84.48 & 89.71 & 91.38& 88.09 \\
$\textup{LoRA}_{r=2}$ &0.33M   & 90.38 & 94.95 & 68.71  & 91.61 & 94.03 & 85.56 & 89.71 & 91.68& 88.33 \\
\textup{AdaLoRA}&0.32M   & \textbf{90.70} & 95.80& 70.04 & 91.78 &\textbf{94.49}& 87.36 & 90.44&91.63& 89.03 \\
\textup{FacT-TT}$_{r=16}$ & 0.06M & 89.91 & 94.72 & 69.69 & 90.76 & 94.18 & 86.28 & 89.71 & 91.70 & 88.36 \\
\textup{FacT-TK}$_{r=32}$ & 0.05M & 90.06 & 95.07 & 69.02 & 91.16 & 94.23 & 87.36 & 89.25 & 91.69 & 88.48\\
$\textup{OFT}_{b=8}$&0.66M   & 89.87& 96.10 & 71.76 & 91.34&93.81& 85.20 & \textbf{90.69}&91.50& 88.78 \\
$\textup{OFT}_{b=4}$&0.33M   & 89.20& 95.18& 70.82 & 90.27 &93.04& 83.39 & 89.46&91.01& 87.80 \\
\midrule
\textup{GOFT} &0.08M   & 90.01 & 96.10& 71.64 & 90.85 &93.52 & 86.28 & 90.44&91.42& 88.78 \\
\textup{qGOFT} &0.33M   & 90.17 & \textbf{96.33}& \textbf{72.08} & 91.34 & 94.03 & \textbf{88.45} & \textbf{90.69} & \textbf{91.72} & \textbf{89.35} \\
\bottomrule
\end{tabular}
\end{small}
\end{center}
\label{tab:glue_res}
\vspace{-0.1in}
\end{table*}

To summarize, qGOFT replaces all $\bm{G}_i$ with $\tilde{\bm{G}_i}$ in each $\bm{P}_r$ of GOFT, and the forward pass is the same as GOFT with an extra regularizer $\sum_i^{d-1} \langle \bm{\alpha}_i, \bm{\beta}_i \rangle ^2$ in the final training objective. qGOFT relaxes the strict orthogonality of Givens rotations with adjustable norms and angular measurements under soft orthogonal constraints. This improves GOFT as well as OFT with a better capability of adapting downstream semantic or conceptual shifts. Moreover, it is worth mentioning that GOFT and qGOFT are both reparameterization tuning methods. Therefore, during inference, we can easily merge the updated weights by multiplying $\bm{P}_r$s to the original weights and conduct the original forwarding, thereby imposing no inference latency and overhead. The overall training procedure is provided in Alg.\ref{alg:training_procedure}, Appendix \ref{apdx:alg}.

\section{Experiments}
In this section, we conduct extensive experiments to evaluate the effectiveness of our methods. We implement GOFT and qGOFT for fine-tuning DeBERTaV3-base \cite{he2021debertav3} and LLaMA2-7B \cite{touvron2023llama}, we also integrate our methods into the PEFT library \cite{peft} \footnote{https://github.com/ArthurLeoM/peft-givens}. Various downstream NLP tasks are applied to fine-tune the PLMs for conducting comparisons between baselines, including natural language understanding \citep[GLUE]{wang2018glue}, instruction following \citep[MMLU]{hendrycks2020measuring} \citep[Vicuna-Eval]{vicuna2023}, and question answering \citep[SQuAD]{rajpurkar2016squad}. We also validate the effectiveness of our method on visual tasks \citep[VTAB-1K]{zhai2019large} by fine-tuning VFMs like ViT-B/16 \cite{dosovitskiy2020image}.

\textbf{Baselines  }
We adopt the following popular and state-of-the-art PEFT methods as our comparison baselines:
\begin{itemize}[leftmargin=*]
    \item \textbf{Full Fine-tuning} \cite{howard2018universal} adjusts all the parameters in the model with gradient updates.
    \item \textbf{BitFit} \cite{zaken2022bitfit} only fine-tunes the bias vectors.
    \item \textbf{H-Adapter} \cite{houlsby2019parameter} applies a two-layer MLP adapter between each self-attention module and FFN module with a subsequent residual connection.
    \item \textbf{P-Adapter} \cite{pfeiffer2020adapterfusion} proposes a more efficient adapter positioned only after each FFN layer and LayerNorm layer.
    \item \textbf{LoRA} \cite{hu2021lora} is the most applied SOTA method for PEFT. LoRA parameterizes the incremental updates of all linear weights by the production of two low-rank learnable matrices and keeps the original parameters frozen.
    \item \textbf{AdaLoRA} \cite{zhang2022adaptive} introduces adaptive parameter budgets to prune the ranks of LoRA in different layers according to sensitivity-based importance scores during the training procedure.
    \item \textbf{FacT} \cite{jie2023fact}, including FacT-TT and FacT-TK, applies shared learnable components to further compress the parameter budgets of low-rank decompositions.
    \item \textbf{OFT} \cite{qiu2023controlling} multiplies a block-diagonal learnable orthogonal matrix with each linear layer to conduct an angular-preserved downstream adaptation.
\end{itemize}
All the mentioned baselines except BitFit have alternative parameter budgets controlled by hyper-parameters of those methods (i.e., the hidden dimension of adapters, the rank of LoRA, and the block size of OFT). However, it is worth mentioning that our methods (i.e., GOFT and qGOFT) do not incorporate hyper-parameters for parameter budgets. For a fair comparison, we mainly focus on the comparisons between methods equipped with similar budgets.

\begin{table*}[t]
\renewcommand\tabcolsep{2.8pt}
\caption{Results of instruction tuning. We report accuracy (\%) for MMLU and average GPT-4-turbo score for Vicuna-Eval. \#Param. denotes the ratio of tunable parameters. The first and second best performing methods are respectively highlighted in \textbf{bold} and \underline{underline}.}
\label{tab:llama_perf}
\begin{center}
\begin{small}
\begin{tabular}{l|c|ccccc|ccccc|c}
\toprule
\multirow{2}{*}{\textbf{Method}}& \multirow{2}{*}{\textbf{\#Param.}}& \multicolumn{5}{c|}{\textbf{MMLU (5-shot)}} & \multicolumn{5}{c|}{\textbf{MMLU (0-shot)}} & \textbf{Vicuna-Eval} \\
 &  & \textup{STEM} & \textup{Social} & \textup{Hum.} & \textup{Other} & \textup{Avg.} & \textup{STEM} & \textup{Social} & \textup{Hum.} & \textup{Other} & \textup{Avg.} & \textup{Score}\\
\midrule
\textup{LLaMA2-7B} & - & 36.80 & 51.42 & 42.76 & 52.10 & 45.49 & 33.31 & 46.78 & 38.76 & 45.04 & 40.79 & 2.66 \\
\textup{LoRA}$_{r=16}$ & 0.25\% & \textbf{37.53} & 50.93 & 42.33 & 52.16 & 45.68 & 34.40 & 45.15 & 38.19 & 45.60 & 40.61 & 5.29\\
\textup{AdaLoRA} & 0.25\% & 36.97 & 52.20 & \underline{42.90} & 52.82 & 45.96 & 34.63 & 48.52 & 40.32 & 47.37 & 42.52 & 5.39 \\
\textup{OFT}$_{b=16}$  & 0.13\% & 37.23 & 51.71 & 42.88 & 52.97 & 45.93 & 34.30 & 49.07 & 40.66 & 48.20 & 42.88 & 5.56 \\
\midrule
\textup{GOFT} & \textbf{0.007\%} & \underline{37.47} & \textbf{52.56} & \textbf{43.27} & \textbf{53.22} & \textbf{46.36} & \underline{35.33} & \textbf{50.18} & \underline{40.77} & \underline{49.32} & \textbf{43.64} & \textbf{5.96} \\
\textup{qGOFT} & \underline{0.03\%} & 37.33 & \underline{52.33} & 42.88 & \underline{53.10} & \underline{46.12} & \textbf{35.50} & \underline{49.33} & \textbf{40.96} & \textbf{49.41} & \underline{43.58} & \underline{5.57} \\

\bottomrule
\end{tabular}
\end{small}
\end{center}
\vskip -0.1in
\end{table*}

\subsection{Natural Language Understanding}
\textbf{Experimental Setting  }
We conduct experiments on the General Language Understanding Evaluation \citep[GLUE]{wang2018glue} benchmark and apply DeBERTaV3-base \cite{he2021debertav3} as our pretrained base model to fine-tune. The model is fine-tuned on the training sets of 8 different tasks from the GLUE benchmark.
The dataset statistics and detailed experimental settings are provided in Appendix~\ref{apdx:GLUE}.

\textbf{Main Results  }
We present the results in Table \ref{tab:glue_res}. Results show that qGOFT consistently outperforms all the baselines under similar parameter budgets. Specifically, we can find GOFT reaches a comparable performance with OFT$_{b=8}$ with 8x parameter reduction and outperforms most baselines as well, revealing the effectiveness of angle-preserving orthogonal transformation for downstream adaptation. However, when shrinking the block size of OFT to 4, the performance will decay intensely. This shows that OFT requires the sacrifice of expressiveness to obtain higher parameter efficiency, whereas GOFT can be as expressive as any orthogonal transformation with an extremely low parameter budget (i.e., 4x lower than other baselines) and obtain compatible results.
Moreover, qGOFT further allows for a more flexible adjustment on angular measurements and weight norms for adapting the slight relative semantic shifts and achieves prominent performances across various tasks.

\subsection{Instruction Tuning}\label{exp:instruction tuning}
\paragraph{Experimental Setting}
In this subsection, we evaluate the natural language generation ability of LLMs tuned by different methods. Specifically, we conduct instruction-tuning for comparisons, which ensures that fine-tuned models follow specific instructions accurately \cite{ouyang2022training}. We select LLaMA2-7B \cite{touvron2023llama} as the base model and conduct fine-tuning on the Alpaca \cite{alpaca} cleaned dataset \footnote{https://huggingface.co/datasets/yahma/alpaca-cleaned} for one epoch.
We apply two evaluation benchmarks including quantitative MMLU \cite{hendrycks2020measuring} and qualitative Vicuna-Eval \cite{vicuna2023}. MMLU consists of 57 different types of multiple-choice tasks and prompts the model to select the correct answer. Vicuna-Eval is an open question-answering benchmark that prompts the model to respond to a predefined set of 80 questions and utilizes GPT-4 series \cite{achiam2023gpt} model to assess the answer qualities, scoring from 1 to 10 with detailed comments.
See Appendix \ref{apdx:llama} for specific hyperparameter settings and task descriptions.

\begin{figure}[t]
\centering
\vspace{-0.1 in}
\includegraphics[width=0.9\linewidth]{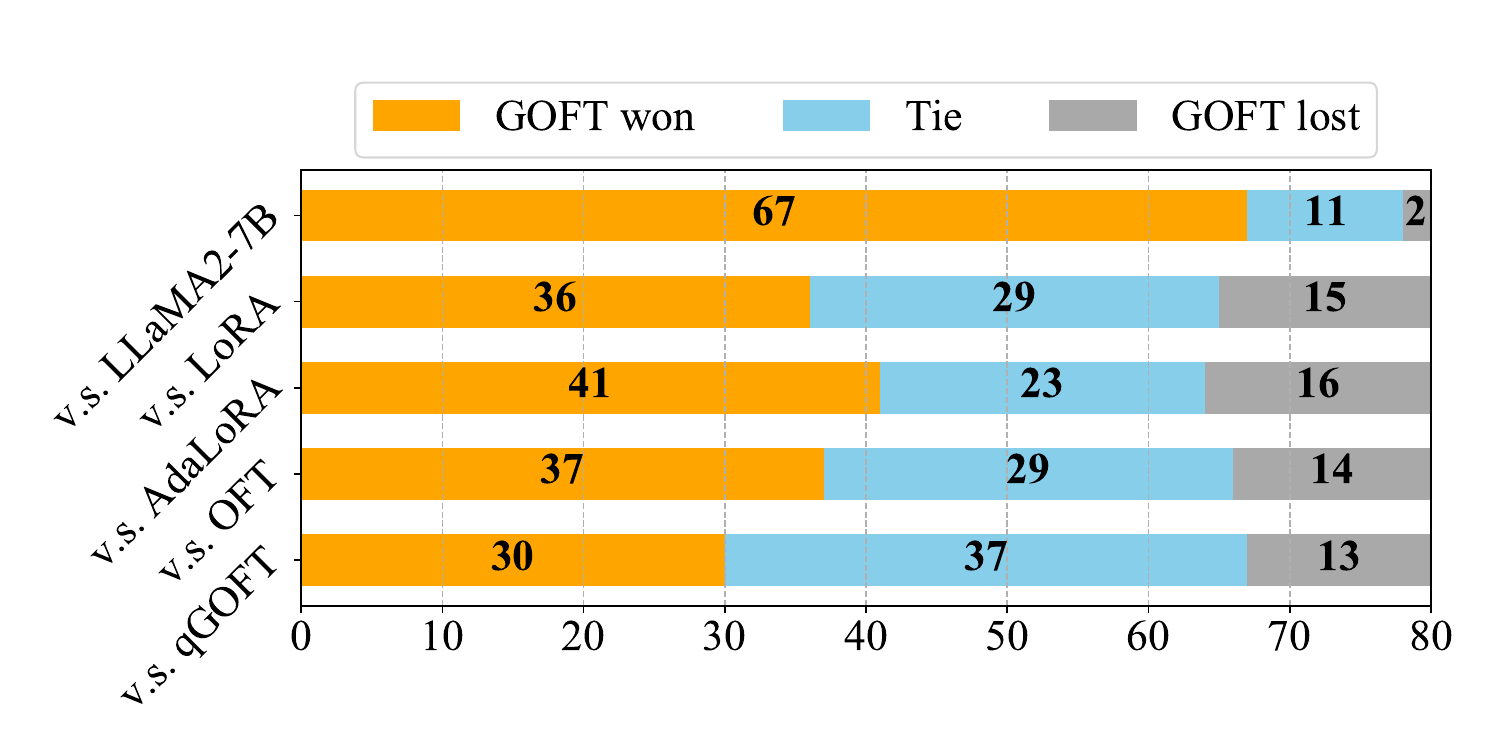} 
\vspace{-10 pt}
\caption{Win rate of GOFT versus other methods on GPT-4-turbo score of Vicuna-Eval benchmark.} 
\label{fig:winrate}
\vspace{-0.1 in}
\end{figure}

\begin{table*}[t]
\renewcommand\tabcolsep{2.1pt}
\caption{Top-1 accuracy (\%) results of VTAB-1K benchmark. The Avg. is obtained by averaging across three categories. \#Param. denotes the number of tunable parameters. The first and second best performing methods are respectively highlighted in \textbf{bold} and \underline{underline}.}
\label{tab:vtab1k_perf}
\begin{center}
\begin{small}
\vspace{-0.15in}
\begin{tabular}{l|c|ccccccc|cccc|cccccccc|c}
\toprule
 & & \multicolumn{7}{c|}{ Natural } & \multicolumn{4}{c|}{ Specialized } & \multicolumn{8}{c|}{ Structured } & \\
 \multirow{5}{*}{\textbf{Methods}} & \multirow{5}{*}{\rotatebox{90}{\# Param.(M)}} & \multirow{5}{*}{\rotatebox{90}{Cifar}} & \multirow{5}{*}{\rotatebox{90}{Caltech101}} & \multirow{5}{*}{\rotatebox{90}{DTD}} & \multirow{5}{*}{\rotatebox{90}{Flower102}} & \multirow{5}{*}{\rotatebox{90}{Pets}} & \multirow{5}{*}{\rotatebox{90}{SVHN}} & \multirow{5}{*}{\rotatebox{90}{Sun397}} & \multirow{5}{*}{\rotatebox{90}{Camelyon}} & \multirow{5}{*}{\rotatebox{90}{EuroSAT}} & \multirow{5}{*}{\rotatebox{90}{Resist45}} & \multirow{5}{*}{\rotatebox{90}{Retinopathy}} & \multirow{5}{*}{\rotatebox{90}{Clevr-Count}} & \multirow{5}{*}{\rotatebox{90}{Clevr-Dist}} & \multirow{5}{*}{\rotatebox{90}{DMLab}} & \multirow{5}{*}{\rotatebox{90}{KITTI-Dist}} & \multirow{5}{*}{\rotatebox{90}{dSpr-Loc}} & \multirow{5}{*}{\rotatebox{90}{dSpr-Ori}} & \multirow{5}{*}{\rotatebox{90}{sNORB-Azim}} & \multirow{5}{*}{\rotatebox{90}{sNORB-Ele}} & \multirow{5}{*}{\rotatebox{90}{Avg.}}  \\
 & & & & & & & & & & & & & & & & & & & & & \\
 & & & & & & & & & & & & & & & & & & & & & \\
 & & & & & & & & & & & & & & & & & & & & & \\
 & & & & & & & & & & & & & & & & & & & & & \\
\midrule 
Full Fine-Tuning & 85.8 & 68.9 & 87.7 & 64.3 & 97.2 & 86.9 & 87.4 & 38.8 & 79.7 & 95.7 & 84.2 & 73.9 & 56.3 & 58.6 & 41.7 & 65.5 & 57.5 & 46.7 & 25.7 & 29.1 & 68.9 \\
Linear & 0 & 64.4 & 85.0 & 63.2 & 97.0 & 86.3 & 36.6 & 51.0 & 78.5 & 87.5 & 68.5 & 74.0 & 34.3 & 30.6 & 33.2 & 55.4 & 12.5 & 20.0 & 9.6 & 19.2 & 57.6 \\ 
BitFit & 0.103 & 72.8 & 87.0 & 59.2 & 97.5 & 85.3 & 59.9 & 51.4 & 78.7 & 91.6 & 72.9 & 69.8 & 61.5 & 55.6 & 32.4 & 55.9 & 66.6 & 40.0 & 15.7 & 25.1 & 65.2 \\
FacT-TT$_{r\leq16}$ & 0.037 & 71.3 & 89.6 & 70.7 & 98.9 & 91.0 & 87.8 & 54.6 & 85.2 & 95.5 & 83.4 & 75.7 & 82.0 & 69.0 & 49.8 & 80.0 & 79.2 & 48.4 & \underline{34.2} & 41.4 & 75.3 \\
FacT-TK$_{r\leq32}$ & 0.069 & 70.6 & 90.6 & 70.8 & 99.1 & 90.7 & \underline{88.6} & 54.1 & 84.8 & 96.2 & 84.5 & 75.7 & 82.6 & 68.2 & 49.8 & 80.7 & 80.8 & 47.4 & 33.2 & 43.0 & 75.6 \\
LoRA$_{r=8}$ & 0.295 & 67.1 & 91.4 & 69.4 & 98.8 & 90.4 & 85.3 & 54.0 & 84.9 & 95.3 & 84.4 & 73.6 & 82.9 & 69.2 & 49.8 & 78.5 & 75.7 & 47.1 & 31.0 & \underline{44.0} & 74.5 \\
OFT$_{b=8}$ & 0.147 & 68.8 & 91.9 & \underline{73.8} & \textbf{99.7} & 92.2 & \textbf{91.8} & 49.2 & \textbf{90.2} & \textbf{100} & \textbf{89.1} & 80.5 & 83.2 & 71.1 & \underline{53.9} & \underline{81.3} & \underline{82.0} & \underline{54.3} & \textbf{34.4} & 43.8 & 78.0 \\
\midrule
GOFT & 0.018 & \underline{75.0} & \textbf{93.9} & 72.3 & \textbf{99.7} & \underline{92.6} & 85.2 & \underline{60.9} & \underline{89.1} & \textbf{100} & \underline{87.9} & \textbf{82.4} & \textbf{84.0} & \textbf{74.2} & \textbf{55.1} & \textbf{82.0} & 80.9 & 52.7 & 32.3 & 43.8 & \underline{78.6} \\
qGOFT & 0.073 & \textbf{81.3} & \textbf{93.9} & \textbf{75.0} & \textbf{99.7} & \textbf{94.5} & 86.9 & \textbf{63.3} & \underline{89.1} & \textbf{100} & 87.5 & \underline{80.9} & \textbf{84.0} & \underline{72.7} & 53.5 & \underline{81.3} & \textbf{83.6} & \textbf{55.1} & 32.3 & \textbf{49.2} & \textbf{79.4} \\

\bottomrule
\end{tabular}
\vspace{-0.1in}
\end{small}
\end{center}
\end{table*}

\textbf{Main Results  }
We present the results alongside the proportion of trainable parameters in Table \ref{tab:llama_perf}, and we provide several specific answers to Vicuna-Eval questions generated by different algorithms in Appendix \ref{apdx:case}.
All the results show that OFT-based methods consistently outperform LoRA-based methods. This can be attributed to the angle-preserving merit of OFT series methods, which better maintain the pretrained concepts and knowledge in LLMs. 
Among all methods, GOFT and qGOFT demonstrate the best performances across quantitative and qualitative metrics. GOFT also shows a prominent winning rate versus all the baselines (see Fig. \ref{fig:winrate}) in the quality of answering open questions. It is especially worth mentioning that GOFT significantly outperforms OFT$_{b=16}$ even with a 16x reduction in parameter costs. This is attributed to GOFT's ability to achieve expressiveness equivalent to any orthogonal transformation while utilizing only $\mathcal{O}(d)$ parameters, while OFT$_{b=16}$ requires sacrificing expressiveness for saving parameter costs. 

An interesting phenomenon is that GOFT performs better than qGOFT in LLaMA instruction tuning, whereas the opposite occurs in the GLUE benchmark. We explain it from two perspectives: (1) the gap between pre-training and downstream tasks, and (2) the impact of model size.
DeBERTa basically applies the encoder architecture and requires an additional trained classifier head for downstream tasks, while LLaMA applies the decoder architecture with an identical pretraining and downstream task, which is to decode and generate. This makes the gap between pretraining and fine-tuning much smaller for LLaMA, which does not require much further adjustment on angular metric spaces or weight norms.
Moreover, \citet{aghajanyan2021intrinsic} have empirically demonstrated that the larger the PLM scale, the fewer adjustments are required to represent a task.
Therefore, in decoder-based and larger PLM (e.g., GPT, LLaMA) instruction tuning, it is more recommended to apply GOFT rather than qGOFT.

\subsection{Question Answering}\label{exp:question answering}

\paragraph{Experimental Setting} In this subsection, we evaluate the proposed algorithms on the question answering benchmark SQuAD v1.1 \cite{rajpurkar2016squad}. 
For comparison, we fine-tuned DeBERTaV3-base with our methods and compared it with other baselines. We report the exact match (EM) accuracy and F1 score as the evaluation metrics. 
For more details, please refer to Appendix \ref{apdx:squad}.

\textbf{Main Results  }
We summarize the results in Table \ref{tab:squad}. The results show that GOFT can reach a comparable performance with OFT with a 4x lower parameter budget. Moreover, qGOFT further enhances GOFT with better capability of downstream adaptation. Specifically, at the same budget levels, qGOFT outperforms existing methods in both EM and F1 evaluation metrics compared with the SOTA baseline. These results demonstrate the superior performance of qGOFT in adapting to QA tasks.

\begin{table}[t]
\caption{Results of SQuADv1.1. We report exact match accuracy (EM) and F1-score. The best results are shown in \textbf{bold}.}
\label{tab:squad}
\begin{center}
\begin{small}
\begin{tabular}{l|c|cc}
\toprule 
 \multirow{2}{*}{\textbf{Method}}&  \multirow{2}{*}{\textbf{\# Param.}} &\multicolumn{2}{c}{SQuADv1.1}  \\
 & & EM &F1\\
\midrule
\textup{Full Fine-Tuning} & 184M & 86.0 & 92.7 \\
\textup{H-Adapter} & 0.31M & 85.3 & 92.1 \\
\textup{P-Adapter} & 0.30M & 85.9 & 92.5 \\
\textup{LoRA}$_{r=2}$ & 0.33M & 86.6 & 92.9 \\
\textup{OFT}$_{b=4}$ & 0.33M   & 85.9 & 92.6 \\
\midrule
\textup{GOFT} &0.08M  &  86.1 & 92.7 \\
\textup{qGOFT} &0.33M   & \textbf{87.1} & \textbf{93.2} \\
\bottomrule
\end{tabular}
\vspace{-0.1in}
\end{small}
\end{center}
\end{table}

\subsection{Adapting Visual Foundation Models}
\textbf{Experimental Setting} Except for natural language tasks, we are also curious about the performance of our methods on visual tasks. Here we conduct experiments on VTAB-1K benchmark \cite{zhai2019large}, which has been intensively applied to evaluate PEFT algorithms for visual tasks. VTAB-1K benchmark consists of 19 different visual datasets, divided into three categories: Natural, Specialized, and Structured. We use the ViT-B/16 \cite{dosovitskiy2020image} pretrained on supervised ImageNet-21K \cite{deng2009imagenet} as the backbone and fine-tune all the linear layers in every transformer block. The detailed experimental settings are provided in Appendix \ref{apdx:vtab}.

\textbf{Main Results} We present the results alongside the size of trainable parameters in Table \ref{tab:vtab1k_perf}. From the results, we can conclude that our methods demonstrate impressive performance compared with the baselines across various visual tasks with a low parameter budget. OFT performs as a SOTA method compared with previous baselines, as it explicitly preserves angular distances in the latent space. Our methods further improve the expressiveness, parameter efficiency, and adaptation capability of block-diagonal OFT, and realize a significant improvement over all the baselines. The results demonstrate the effectiveness of our method in fine-tuning vision transformers.

\subsection{Ablation Studies}\label{exp:ablation}

\begin{figure}[t]
\centering
\includegraphics[width=1.05\linewidth]{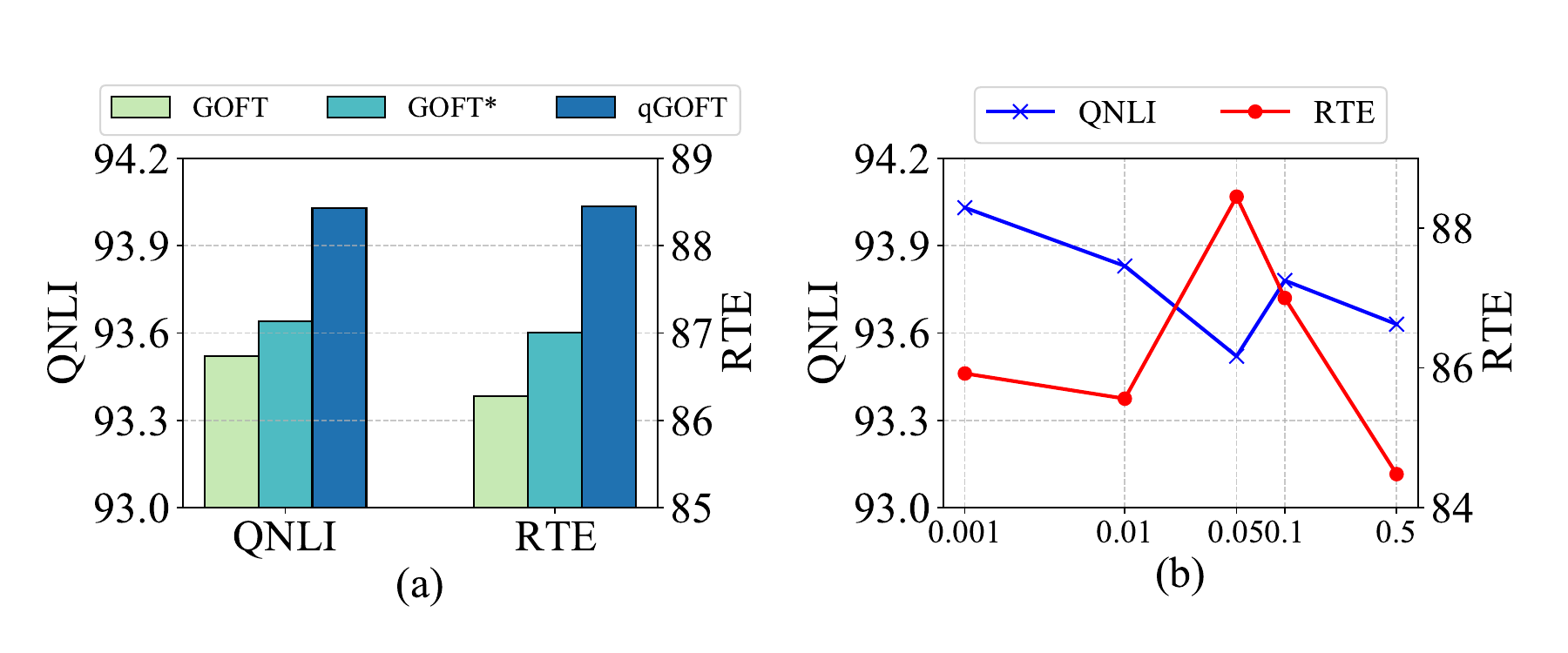} 
\vspace{-0.3in}
\caption{Ablation Studies: (a) Comparisons of GOFT, qGOFT, and GOFT with only norm adjustment (GOFT*). (b) Varying orthogonal regularization strength $\lambda$ in qGOFT.} 
\label{fig:abl}
\vspace{-0.2in}
\end{figure}

\textbf{Introducing Different Flexibility into GOFT  }\label{exp:ablation Different Strategies of Givens}
We conduct experiments of introducing different levels of flexibility into GOFT on QNLI and RTE datasets. Specifically, GOFT* denotes that only norm adjustments are conducted by a learnable diagonal matrix multiplying with the learned Givens rotation, which is still an angular-preserving method, while qGOFT incorporates both norm and angular adjustments. The results are shown in Fig.\ref{fig:abl}(a). We can see that GOFT* can improve the performances to some extent, demonstrating the effectiveness of adapting the attention distribution shifts in downstream tasks, which can be depicted by norm adjustments. While qGOFT further enhances GOFT* by a significant margin and illustrates the importance of adapting relative semantic shifts from downstream corpora.

\textbf{Effect of Orthogonal Regularization Strength $\lambda$ on qGOFT Performances}\label{exp:ablation different reg}
We select various values of $\lambda$ in qGOFT ranging from \{0.001, 0.01, 0.05, 0.1, 0.5\} and conduct experiments on QNLI and RTE datasets.
The results are illustrated in Fig.\ref{fig:abl}(b). We can observe that qGOFT cannot perform quite well when taking a too large $\lambda$ (i.e., 0.5), which is restricting relatively strong orthogonality, as the adjustments on relative semantic angular distances are inadequate. However, this does not mean the smaller the $\lambda$, the better the performance. Though QNLI shows the best performance at the minimum $\lambda=$0.001, RTE reaches the highest accuracy only at $\lambda=$ 0.05. This indicates that the selection of $\lambda$ depends on the downstream task and requires trial-and-error for selecting an appropriate setting.

\section{Conclusions and Future Works}
In this work, we introduce a novel and generic parameter-efficient fine-tuning algorithm for pretrained models based on quasi-Givens rotations, dubbed quasi-Givens Orthogonal Fine-Tuning (qGOFT). 
qGOFT improves orthogonal fine-tuning by tackling two significant problems: 1) low parameter efficiency, where we reduce the parameter costs from $\mathcal{O}(d^2)$ to $\mathcal{O}(d)$ with rigorously equivalent expressiveness to OFT in $SO(d)$; 2) limited adaptation capability of downstream semantic shifts, where we enable flexible norms and relative angular adjustments under soft orthogonal regularization.
We evaluate the effectiveness of qGOFT with various NLP and vision tasks on different pretrained models, including NLU and QA tasks on DeBERTaV3, NLG on LLaMA2, and image classification tasks on ViT-B/16. All experiments demonstrate the empirical effectiveness and superiority of qGOFT as a generic PEFT framework. 
We have also integrated it into the peft library to contribute to the open-source LLM community.

Despite the promising results obtained in our work, it is important to acknowledge its limitations.
The training runtime overhead is slightly larger than the original OFT due to the product of $\mathcal{O}(\log d)$ Givens rotation matrices. In essence, this is a trade-off between time complexity and parameter complexity.
Fortunately, there are some works \cite{chen2022pixelated} focusing on accelerating multiple sparse matrix multiplications via the approximation of the first-order Taylor expansion, converting time-consuming multiplicative operations to fast additions.
We regard this as an important future work and expect to present a more efficient and effective version of qGOFT.


\section*{Impact Statement}
This paper presents work whose goal is to advance the field of Machine Learning. There are many potential societal consequences of our work, none of which we feel must be specifically highlighted here.

\section*{Acknowledgements}
This work is supported by the National Natural Science Foundation of China (No.U23A20468). We would also like to sincerely thank Prof. Yasha Wang for productive discussions and helpful suggestions.

\bibliography{example_paper}
\bibliographystyle{icml2024}

\newpage
\appendix
\onecolumn

\section*{Organization of the Appdendix}
Appendix \ref{apdx:figs} presents the illustrative examples of theoretical results in Section \ref{sec:method_1}.

Appendix \ref{apdx:impl_key} presents some implement keys of (q)GOFT.

Appendix \ref{apdx:alg} presents the algorithm framework of qGOFT.

Appendix \ref{apdx:GLUE} presents experimental details of the GLUE benchmark, including implementation details and dataset statistics.

Appendix \ref{apdx:llama} presents experimental details of LLaMA-2-7B instruction tuning and introductions to the evaluation benchmark.

Appendix \ref{apdx:squad} presents experimental details of the SQuAD v1.1 benchmark.

Appendix \ref{apdx:vtab} presents experimental details of the VTAB-1K benchmark.

Appendix \ref{apdx:overhead} presents training time comparisons between our methods and selected baselines.

Appendix \ref{apdx:case} presents four case studies of answers to the questions in the Vicuna-Eval benchmark given by models fine-tuned with different methods.

\section{Illustrative Examples of Theoretical Results} 
\label{apdx:figs}
\begin{figure*}[h]
    \centering
    \includegraphics[width=0.9\linewidth]{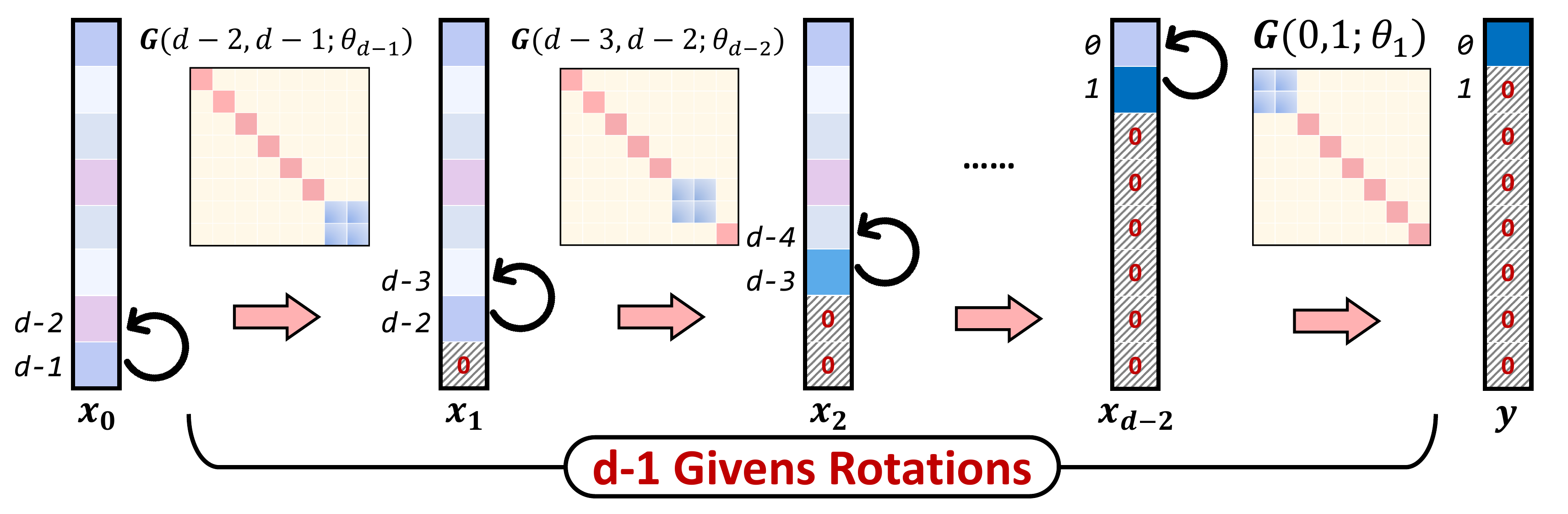}
    \caption{The illustrative example (i.e., rotating procedure) of our proof.}
    \label{fig:proof}
    \vspace{-0.1in}
\end{figure*}

\begin{figure*}[h]
    \centering
    \includegraphics[width=0.8\linewidth]{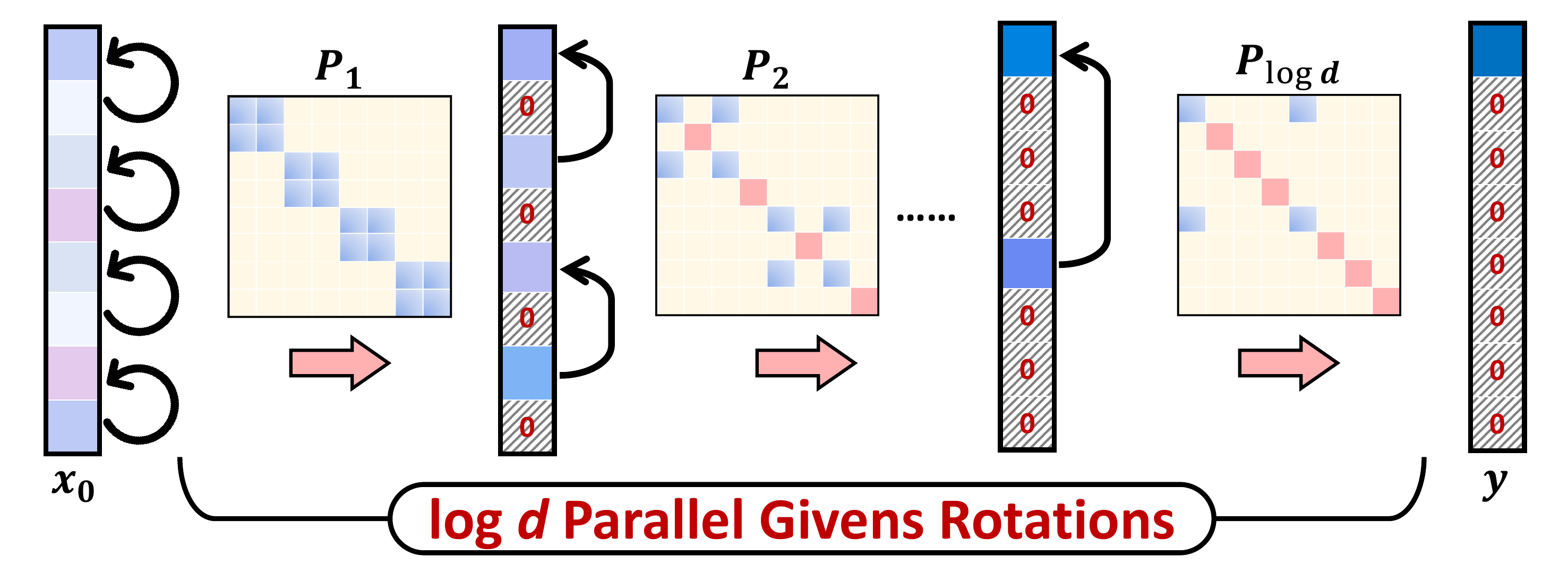}
    \caption{The illustrative example of parallel Givens rotation.}
    \label{fig:parallel}
    \vspace{-0.1in}
\end{figure*}

\addcontentsline{toc}{section}{Illustrative Examples of Theoretical Results}
We present the illustrative rotation strategy given in the proof of Theorem \ref{theo:givens} with Fig. \ref{fig:proof}, and we also present the illustrative parallel rotation idea introduced in Section \ref{sec:method_1} with Fig. \ref{fig:parallel}. See main texts for detailed explanations.

\section{Some Key Points for Implementation} \label{apdx:impl_key}
When implementing GOFT and qGOFT, there are some key points requiring attention:
\begin{itemize}
    \item \textbf{Identity Initialization:} There is a consensus in PEFT that the fine-tuned model will not drift too far from the original parameters. Hence, conventional manners, such as LoRA-series methods, apply zero initialization of the additive delta weights. Therefore, we also follow this paradigm and use an identity matrix as the initialization of every (quasi-)Givens rotation (i.e., zero initialization of $\bm{\theta}$ in GOFT and identity initialization of $\bm{P}_r$ in qGOFT). 
    \item \textbf{Sparse Matrix Multiplications:} As $\bm{P}_r$s are all sparse matrices, applying sparse matrix multiplications can reduce GPU memory costs. However, this is not essential and will not greatly influence the experimental results.
    \item \textbf{No Inference Overhead:} Our methods are basically reparameterization tuning methods. Therefore, during inference, we can easily merge the delta weights by multiplying $\bm{P}_r$s to the original weights and conduct the original forwarding, thereby imposing no inference latency and overhead.
\end{itemize}

\section{qGOFT Algorithm Framework} \label{apdx:alg}
\begin{algorithm}[h]
    \caption{The fine-tuning and testing procedure of a pre-trained model with (q)GOFT.}
    \label{alg:training_procedure}
\begin{algorithmic}[1]
    \STATE {\bfseries Input:} Training dataset $\mathcal{D}$, Frozen Pretrained model $\mathcal{M}$, Tunable quasi-Givens Rotation Matrices $\tilde{\bm{G}}_i^n = [\bm{\alpha}_i^n, \bm{\beta}_i^n]$ denoting the $i$-th quasi-Givens Rotation of the $n$-th linear layer.
    \STATE {\bfseries Training Stage:}
        \STATE Initialization: $\tilde{\bm{G}}_i^n := \bm{I}$.
        \FOR{each iteration}
            \STATE Randomly draw a mini-batch of samples from the training set $\mathcal{D}$;
            \STATE Parallel rotating the weight matrix $\bm{W}^n$ of the fine-tuned linear layers in $\mathcal{M}$ using Eq.(\ref{eq:parallel})
            \STATE Conduct forward pass using Eq.(\ref{eq:fw_pass}).
            \STATE Calculate loss function $\mathcal{L} = \mathcal{L}_{tr} + \lambda \sum_i \sum_n \langle \bm{\alpha}_i^n, \bm{\beta}_i^n \rangle ^2$;
            \STATE Update quasi-Givens rotation matrices $\tilde{\bm{G}}_i^n$ with $\nabla\mathcal{L}$.
        \ENDFOR
    \STATE {\bfseries Test Stage:}
        \STATE Merge Delta Weights: Update all the tuned linear layers $\bm{W}^n$ in $\mathcal{M}$ with $\bm{W}^{n}_* = (\prod_{r} \bm{P}_r^n ) \bm{W}^n$.
        \FOR{each sample $\bm{x}$ in test set}
        \STATE Inference using original forward function with $\bm{W}^n_*$.
        \ENDFOR
\end{algorithmic}
\end{algorithm}

\section{Experimental Details of GLUE Benchmark} \label{apdx:GLUE}
\subsection{Implementation Details}
We use PyTorch library to implement all the algorithms based on the open-source HuggingFace transformers \cite{wolf2019huggingface} codebase.
For all our experiments on GLUE benchmark, we follow the model architecture setting of \cite{zhang2022adaptive} and only tune the model-agnostic hyperparameters (e.g., learning rate, training epochs, regularization strength, etc.). For all reparameterization tuning methods including LoRA, AdaLoRA, OFT, GOFT and qGOFT, we fine-tuned all the linear layers in every transformer block. The experiments are conducted on a single NVIDIA-A100-80GB GPU or distributedly on a maximum of 4 NVIDIA-RTX3090-24GB GPUs. The specific tuned hyperparameters used in our experiments are presented in Table \ref{tab:hyp_glue}.

\begin{table}[h]
    \centering
    \caption{Hyperparameter settings in GLUE benchmark experiments.}
    \vskip 0.1in
    \label{tab:hyp_glue}
    \begin{tabular}{c|c|cccccccc}
    \toprule
    Method & Dataset & MNLI & SST-2 &  CoLA &  QQP & QNLI & RTE & MRPC & STS-B \\
    \midrule
    \multirow{3}{*}{\begin{tabular}{c} 
    DeBERTaV3-base \\
    GOFT
    \end{tabular}} & Batch Size & 32 & 32 & 32 & 32 & 32 & 32 & 32 & 32 \\
    & \# Epochs & 6 & 8 & 20 & 5 & 5 & 30 & 30 & 20 \\
     & Learning Rate & 5e-4 & 8e-4 & 1.6e-3 & 5e-4 & 1.2e-3 & 2e-3 & 8e-4 & 7e-4 \\
    \midrule \multirow{4}{*}{\begin{tabular}{c} 
    DeBERTaV3-base \\
    qGOFT
    \end{tabular}} & Batch Size & 32 & 32 & 32 & 32 & 32 & 32 & 32 & 32 \\
     & \# Epochs & 6 & 8 & 20 & 5 & 5 & 30 & 30 & 20 \\
     & Learning Rate & 6e-4 & 8e-4 & 8e-4 & 1.5e-3 & 1.2e-3 & 2e-3 & 1.2e-3 & 5e-4 \\
     & Reg. Strength $\lambda$ & 0.001 & 0.01 & 0.05 & 0.001 & 0.001 & 0.05 & 0.01 & 0.001 \\
    \bottomrule
    \end{tabular}
\end{table}
\vspace{-0.1in}
\subsection{Dataset Statistics}
We present the detailed dataset statistics of GLUE benchmark \cite{wang2018glue} in Table \ref{tab:glue_stat}.

\begin{table*}[h]
    \caption{Dataset Statistics of GLUE benchmark}
    \vskip 0.1in
    \label{tab:glue_stat}
    \centering
    \begin{tabular}{l|l|c|c|c|c}
\toprule
\textbf{Dataset} & \textbf{Task} & \textbf{\#Train} & \textbf{\#Dev}  & \textbf{\#Label} & \textbf{Metrics} \\
\midrule 
\multicolumn{6}{c}{ Single-Sentence Classification } \\
\hline CoLA & Acceptability & 8.5k & 1k & 2 & Matthews Corr \\
\hline SST & Sentiment & 67k & 872  & 2 & Accuracy \\
\midrule
\multicolumn{6}{c}{ Pairwise Text Classification } \\
\hline MNLI & NLI & 393k & 20k & 3 & Accuracy \\
\hline RTE & NLI & 2.5k & 276 & 2 & Accuracy \\
\hline QQP & Paraphrase & 364k & 40k & 2 & Accuracy \\
\hline MRPC & Paraphrase & 3.7k & 408 & 2 & Accuracy \\
\hline QNLI & QA/NLI & 108k & 5.7k & 2 & Accuracy \\
\midrule
\multicolumn{6}{c}{ Text Similarity } \\
\hline STS-B & Similarity & 7k & 1.5k & 1 & Pearson Corr \\
\bottomrule
\end{tabular}
\end{table*}

\section{Experimental Details of LLaMA-2 Instruction Tuning} \label{apdx:llama}
\subsection{Implementation Details}
We use PyTorch library to implement all the algorithms based on the open-source HuggingFace transformers \cite{wolf2019huggingface} codebase. For the fine-tuning algorithms, we implement the baselines and our methods using the PEFT library \cite{peft}. For the training and evaluation framework, we apply the LLaMA-Factory open-source library \cite{llama-factory}. We tuned all the 'q\_proj', 'v\_proj', 'k\_proj', and 'o\_proj' linear modules in every transformer block of LLaMA-2 model for all the compared baselines. For the instruction tuning, we apply the alpaca template with the Alpaca-cleaned-51K dataset \footnote{\url{https://huggingface.co/datasets/yahma/alpaca-cleaned}} to construct our training set. The experiments are conducted distributively on two NVIDIA-A100-80GB GPUs with deepspeed \cite{rasley2020deepspeed} parallel framework.
The detailed hyperparameters are provided in Table \ref{tab:llama_hyp}.

\begin{table}[h]
\caption{Hyperparamters Used in LLaMA-2-7B fine-tuning.}
\vskip 0.1in
    \label{tab:llama_hyp}
    \centering
    \begin{tabular}{l|cccccc}
        \toprule
        \textbf{Method} & batch size & grad. acc. step & learning rate & warm-up steps & lr-scheduler & Reg. Strength $\lambda$ \\
        \midrule
        LoRA$_{r=16}$ & 32 & 2 & 5e-5 & 50 & cosine & -\\
        OFT$_{b=16}$ & 32 & 2 & 5e-5 & 50 & cosine & -\\
        GOFT & 32 & 2 & 1e-4 & 50 & cosine & -\\
        qGOFT & 32 & 2 & 1e-4 & 50 & cosine & 0.01\\
        \bottomrule
    \end{tabular}
\end{table}

\subsection{Details of Alpaca Cleaned Dataset}
The original Alpaca \cite{alpaca} is a dataset of 52,000 instructions and demonstrations generated by OpenAI's text-davinci-003 engine. This instruction data can be used to conduct instruction-tuning for language models and make the language model follow instruction better.
Alpaca-cleaned-51K is a cleaned version of the original Alpaca Dataset released by Stanford, which mainly fixed the following issue in the original instruction sets, including hallucinations, merged instructions, empty outputs, wrong answers, etc.
The cleaned version can improve the performance of natural language processing models trained on this data and is suitable for instruction tuning.

\subsection{Details of Vicuna-Eval Benchmark}
Vicuna-Eval consists of 80 natural language questions comprising eight problem categories such as Fermi problems, roleplay scenarios, and coding/math tasks, to test various aspects of a model's instruction-following and question-answering performance. We then utilize LLM-judge open-source library\footnote{\url{https://github.com/lm-sys/FastChat/tree/main/fastchat/llm_judge}} to apply GPT-4-turbo to rate the quality of answers from different fine-tuned models based on helpfulness, relevance, accuracy, and detail, score ranging from 1 to 10. \citet{zheng2023judging} claim that GPT-4 can produce not only relatively consistent scores but also detailed explanations on why such scores are given. We provide the judging prompt as follows, which is consistent with \cite{vicuna2023, zheng2023judging}.

\begin{center}
    \fcolorbox{black}{gray!10}{
    \parbox{0.9\linewidth}{
    \ttfamily [System] 
    
Please act as an impartial judge and evaluate the quality of the response provided by an AI assistant to the user question displayed below. Your evaluation should consider factors such as the helpfulness, relevance, accuracy, depth, creativity, and level of detail of the response. Begin your evaluation by providing a short explanation. Be as objective as possible. After providing your explanation, please rate the response on a scale of 1 to 10 by strictly following this format: "[[rating]]", for example: "Rating: [[5]]".

[Question] 

\{question\} 

[The Start of Assistant’s Answer] 

\{answer\} 

[The End of Assistant’s Answer] 
    }
    }
\end{center}

\section{Experimental Details of SQuADv1.1 Benchmark} \label{apdx:squad}
\subsection{Implementation Details}
We use PyTorch library to implement all the algorithms based on the open-source HuggingFace transformers \cite{wolf2019huggingface} codebase.
For all our experiments on SQuADv1.1 benchmark, we follow the model architecture setting of \cite{zhang2022adaptive} and only tune the model-agnostic hyperparameters. We also fine-tuned all the linear layers in every transformer block for GOFT and qGOFT. The experiments are conducted on a single NVIDIA-A100-80GB GPU. The specific tuned hyperparameters used in our experiments are presented in Table \ref{tab:hyp_squad}.

\begin{table}[h]
\caption{Hyperparamters Used in SQuADv1.1 Experiments.}
\vskip 0.1in
    \label{tab:hyp_squad}
    \centering
    \begin{tabular}{l|cccccc}
        \toprule
        \textbf{Method} & batch size  & epochs & learning rate & Reg. Strength $\lambda$ \\
        \midrule
        GOFT & 32 & 10 & 8e-4 & - \\
        qGOFT & 32 & 10 & 8e-4 & 0.001 \\
        \bottomrule
    \end{tabular}
\end{table}

\subsection{Benchmark Details}
Stanford Question Answering Dataset (SQuAD) v1.1  \cite{rajpurkar2016squad} is a question-answering dataset, consisting of questions posed by crowdworkers on a set of Wikipedia articles, where the answer to every question is a segment of text, or span, from the corresponding reading passage. SQuADv1.1 benchmark formulates a sequence labeling problem of QA pairs, where the goal is to predict whether a token is the start or the end of an answer span. Specifically, SQuADv1.1 consists of 87,599 training samples and 10,570 validation samples.

\newpage

\section{Experimental Details of VTAB-1K Benchmark} \label{apdx:vtab}
\subsection{Implementation Details}
We use PyTorch library to implement all the algorithms based on the open-source HuggingFace transformers \cite{wolf2019huggingface} codebase.
For all our experiments on VTAB-1K benchmark, we follow the model architecture setting and training framework of \citet{jie2023fact} \footnote{\url{https://github.com/JieShibo/PETL-ViT/tree/main/FacT}}. We also fine-tuned all the linear layers in every transformer block for GOFT and qGOFT. The experiments are conducted on a single NVIDIA-RTX3090-24GB GPU. The specific tuned hyperparameters used in our experiments are presented in Table \ref{tab:hyp_vtab}.

\begin{table}[h]
\caption{Hyperparamters Used in VTAB-1K Experiments.}
\vskip 0.1in
    \label{tab:hyp_vtab}
    \centering
    \begin{tabular}{l|cccccc}
        \toprule
        \textbf{Method} & batch size  & epochs & learning rate & weight decay & Reg. Strength $\lambda$ \\
        \midrule
        GOFT & 64 & 100 & 1e-3 & 1e-4 & - \\
        qGOFT & 64 & 100 & 1e-3 & 1e-4 & 0.001 \\
        \bottomrule
    \end{tabular}
\end{table}

\subsection{Dataset Description}
Visual Task Adaptation Benchamark (VTAB-1K)\cite{zhai2019large} consists of 19 image classification tasks that are divided into three categories: natural, specialized, and structured. Natural image tasks involve photographs of the natural environment taken with regular cameras, depicting generic objects, fine-grained classes, or abstract concepts. Specialized tasks utilize images taken with specialized equipment like medical imaging devices or remote sensing technology. Structured tasks often arise from artificial settings aimed at analyzing specific changes between images, such as estimating object distances in a 3D scene (e.g., DMLab), object counting (e.g., CLEVR), or orientation detection (e.g., dSprites for disentangled representations). In VTAB-1k, each dataset provides 800 labeled training set samples, a subset of their original training set. We use them to fine-tune our base model and the Top-1 classification accuracy on their respective original test set is used as the performance measure.

\section{Training Time Overhead Comparisons} 
\label{apdx:overhead}
We take experiments on QNLI of GLUE benchmark as an example to conduct runtime comparisons between our methods and selected baselines. The experiment is conducted on two NVIDIA-RTX3090-24GB GPUs, and the hyperparameter settings are the same as introduced in Appendix \ref{apdx:GLUE}. We report the averaged runtime (seconds) per batch of different methods in Table \ref{tab:time_stat} for comparisons.

\begin{table}[h]
\caption{Training time (seconds) per batch using different PEFT methods to fine-tune QNLI dataset.}
\vskip 0.1in
    \label{tab:time_stat}
    \centering
    \begin{tabular}{l|ccccc}
        \toprule
        \textbf{Method} & LoRA & AdaLoRA & OFT & GOFT & qGOFT \\
        \midrule
        Time(s)/Batch & 1.12 & 1.43 & 1.80 & 6.00 & 6.27 \\
        \bottomrule
    \end{tabular}
\end{table}

We can see that the overall runtime of (q)GOFT is around 3.3-3.5x of OFT. The main overhead comes from the $\mathcal{O}(\log d)$ sparse matrix multiplications (spmms) of each layer in our methods, just as we discussed in the limitations. It is also worth mentioning that the runtime is not strictly $\log d$ times of OFT, because spmm can be conducted faster than dense matrix multiplication. In a nutshell, this is basically a trade-off between time complexity and parameter complexity. We traded 3.3-3.5x of time costs for $\mathcal{O}(d^2)$ to $\mathcal{O}(d)$ parameter saving with the same expressiveness, and we think the current training runtime is still acceptable. Moreover, the sequential spmms can be further optimized by distributive GPU calculations to be supported by CUDA distributive hardware. In our future work, as mentioned in the paper, we will focus on reducing the time complexity of our methods via optimizing algorithms and distributed calculation frameworks.

\newpage
\section{Case Study of Vicuna-Eval Benchmark} \label{apdx:case}
\subsection{Case 1: Generic QA}
The first case is about generic QA asking \textit{How can I improve my time management skills?} The highest score is 9 obtained by GOFT, with no negative comments given by GPT-4. LoRA, AdaLoRA and qGOFT obtain 8 points, and OFT obtains 7. Among those methods, LoRA presents repetitive points of view, and qGOFT and OFT may lack a little depth. The original LLaMA-2-7B fails to provide valuable answers. Refer to Table \ref{tab:Vicuna question1} for the detailed answers and GPT-4-turbo judgements.

\begin{longtable}{|p{1.6cm}|p{14cm}|}
\caption{Vicuna question \#.1} \\
 \toprule
    \textbf{Question}&\textbf{How can I improve my time management skills?} \\
		\midrule
		 \multicolumn{2}{|c|}{\textbf{Method: GOFT, Score: 9}}  \\
		\midrule
  \endfirsthead
\multicolumn{2}{c}%
{{\bfseries \tablename\ \thetable{} -- continued from previous page}} \\
\endhead
\bottomrule
\endlastfoot
         \textbf{Answer} 
        &{Time management is a skill that can be improved through several techniques. Here are some tips that may help you to better manage your time and be more productive:\newline  1. Prioritize your tasks: Start by prioritizing your tasks and focusing on the most important ones. Make a list of everything you need to do and decide which tasks are the most important and need to be completed first.\newline  2. Set a schedule: Plan your day or week ahead of time. Set aside time for work, leisure, and other activities. Schedule your tasks in your calendar, and stick to your schedule as much as possible.\newline  3. Break down large tasks: If you have a big task ahead of you, break it down into smaller tasks and assign a time limit for each one. This will help you to stay on track and be more productive.\newline  4. Stay organized: Keep your workspace clean and organized. Use a to-do list, calendar, or other organizational tools to help you stay on track.\newline  5. Avoid distractions: Distractions can be a major time waster. Avoid checking social media, answering phone calls, or taking breaks until you have completed your most important tasks.\newline  6. Learn to say "no": It can be hard to say no to people and commitments, but it's important to be able to say no to things that will take up your time and prevent you from completing your tasks. Learn to prioritize your time and say no to things that don't align with your goals.\newline  7. Take breaks: Taking short breaks during your workday can help you to stay focused and avoid burnout. Get up from your desk and take a walk, or do something else that refreshes your mind.\newline  8. Get enough sleep: Getting enough sleep is important for managing your time. A well-rested mind is more productive and better able to focus.\newline  9. Learn to delegate: If you're feeling overwhelmed, consider delegating tasks to other people. Delegating tasks can help you to manage your time more effectively and avoid burnout.\newline  10. Seek help: If you're having trouble managing your time, seek help from a time management expert or coach. They can help you to identify areas where you need improvement and provide guidance and support.}\\
        \cline{1-2}
		\textbf{Judgement} &{The AI assistant's response to the question about improving time management skills is comprehensive, relevant, and practical. The answer provides a list of actionable tips that cover a wide range of strategies, from prioritizing tasks and setting a schedule to avoiding distractions and learning to delegate. \textbf{Each point is explained clearly and concisely, making it easy for the user to understand and apply the advice in their daily life.}\newline  The response also touches on the importance of self-care (such as taking breaks and getting enough sleep) in maintaining productivity, which is an essential aspect of time management that is often overlooked. The suggestion to seek professional help if needed is a valuable addition, as it acknowledges that some individuals may require more personalized assistance.\newline The answer does not include unnecessary jargon or complex concepts, which makes it accessible to a broad audience. \textbf{It also avoids being overly simplistic, providing enough detail to be genuinely helpful without overwhelming the reader.}\newline  Overall, the AI's response is well-structured, informative, and likely to be very helpful to someone looking to improve their time management skills. \textbf{It does not appear to contain any inaccuracies or misleading information.} }
        \\\hline
        \newpage
		\midrule
        \multicolumn{2}{|c|}{\textbf{Method: qGOFT, Score: 8}}  \\
		\midrule
        Answer       &{Time management is a skill that can be improved through regular practice and a commitment to developing and honing these skills. Here are some ways you can improve your time management skills:\newline 1. Prioritize tasks: The first step in time management is to prioritize your tasks. Make a list of everything you need to do, and rank the tasks by priority. Focus on the most important tasks first, and allocate enough time to each task to ensure you are giving it the attention it needs.\newline 2. Plan your day: Create a daily plan that outlines your tasks for the day, including any meetings, deadlines, or appointments. This will help you stay organized and on track throughout the day.\newline 3. Use a time management tool: There are many time management tools available that can help you track your time, schedule tasks, and stay organized. Some popular options include Google Calendar, Asana, or Trello.\newline 4. Set goals and deadlines: Set goals and deadlines for your tasks to ensure you stay on track. This will help you stay motivated and focused, and will also help you stay organized and avoid procrastination.\newline 5. Manage distractions: Distractions can be a major time waster. Learn to manage them by turning off notifications, avoiding social media, and finding ways to focus on your tasks.\newline 6. Take breaks: It is important to take breaks throughout the day to recharge your mind and body. Take a few minutes to relax and refocus, this will help you be more productive when you return to your tasks.\newline 7. Learn to say no: Learn to say no to tasks that are not a priority, or that will take too much time. By saying no to unimportant tasks, you will be able to focus on the tasks that are truly important to you.\newline By practicing these time management skills, you will be able to improve your productivity, and better manage your time and tasks. Remember, time management is a skill that takes practice and dedication, but the rewards are well worth the effort.}\\
        \cline{1-2}
       Judgement&{\textbf{The AI assistant's response to the question about improving time management skills is helpful, relevant, and accurate.} The suggestions provided are commonly recommended strategies for enhancing time management, such as creating a daily schedule, prioritizing tasks, setting deadlines, taking breaks, planning ahead, avoiding distractions, and using time management tools. \textbf{The response is clear and easy to understand}, offering practical advice that can be implemented by the user.\newline \textbf{The answer could be improved by providing more depth in some areas}, such as explaining how to prioritize tasks effectively or recommending specific time management tools and how to use them. Additionally, the response could include creative tips or lesser-known strategies for managing time more efficiently. However, \textbf{the advice given is solid and covers the basics well.}}\\
       		\midrule
                \newpage
                \midrule
        \multicolumn{2}{|c|}{\textbf{Method: OFT, Score: 7}}  \\
		\midrule
        \textbf{Answer }&{Improving your time management skills can be a challenge, but it is definitely possible with the right approach. Here are some tips that can help you manage your time more effectively:\newline 1. Prioritize: Start by prioritizing your tasks and setting aside time for each one. This will help you focus on the most important tasks first and ensure that you are using your time wisely.\newline 2. Plan ahead: Plan your day or week ahead of time, including any meetings, appointments, or deadlines. This will help you stay organized and avoid last-minute rushes.\newline 3. Set goals: Set specific, achievable goals for yourself each day, week, or month. This will help you stay motivated and on track.\newline 4. Stay organized: Keep your workspace organized and decluttered. This will help you find what you need quickly and avoid wasting time looking for things.\newline 5. Say no: Learn to say no to requests or tasks that you don't have time for. This will help you focus on the most important tasks and avoid feeling overwhelmed.\newline 6. Take breaks: Take regular breaks throughout the day to recharge and refresh your mind. This will help you stay focused and productive.\newline 7. Manage distractions: Learn to manage distractions, such as social media, email, or phone calls. This will help you stay focused on your tasks and avoid wasting time.\newline8. Be realistic: Be realistic about how much you can accomplish in a day or week. This will help you avoid feeling overwhelmed and stressed.\newline 9. Be flexible: Be flexible and willing to adjust your plans as needed. This will help you stay on track and avoid getting stuck in a rut.\newline10. Seek help: Don't be afraid to seek help or advice from others if you are struggling with time management. There are many resources available, such as books, online courses, or coaching.\newline By following these tips, you can improve your time management skills and become more efficient and productive. Remember, it takes time and practice to develop good habits, but the effort will be worth it in the long run.}\\
        \cline{1-2}
        \textbf{Judgement} &{\textbf{The AI assistant's response to the question about improving time management skills is comprehensive, relevant, and practical.} The answer provides a list of actionable tips that are commonly recommended for better time management, such as prioritizing tasks, planning ahead, setting goals, staying organized, learning to say no, taking breaks, managing distractions, being realistic, being flexible, and seeking help. Each point is briefly explained, giving the user a clear understanding of how to apply these strategies in their daily life.\newline The response is well-structured, with each tip forming a separate bullet point, which makes it easy to read and digest. The advice is also general enough to be applicable to a wide audience, regardless of their specific profession or lifestyle. Additionally, the assistant encourages the user to be patient and persistent, acknowledging that developing good time management habits takes time and practice.\newline \textbf{However, the response could be improved by providing more depth in some of the tips}, such as offering specific techniques for prioritizing tasks or examples of how to set achievable goals. Including information about common time management methodologies like the Eisenhower Matrix, Pomodoro Technique, or time blocking could also enhance the quality of the advice. Furthermore, the assistant could have mentioned the importance of self-reflection and regular review of one's time management practices to ensure continuous improvement.\newline Overall, the response is helpful, accurate, and provides a solid foundation for someone looking to improve their time management skills. \textbf{It lacks some depth and creativity that could make the advice more impactful}, but it is a good starting point for most individuals.}\\
        \hline
        \newpage
		\midrule
        \multicolumn{2}{|c|}{\textbf{Method: AdaLoRA, Score: 8}}  \\
		\midrule
 Answer &{Time management is a skill that can be improved through a combination of self-awareness, planning, and execution. Here are some strategies that can help you improve your time management skills:\newline 1. Prioritize: Identify the most important tasks and prioritize them accordingly. This will help you focus on the tasks that will have the greatest impact on your goals.\newline 2. Set goals: Set specific, measurable, and achievable goals for yourself. This will help you stay focused and motivated to complete your tasks.\newline 3. Plan ahead: Plan your day or week in advance, including the time you will spend on each task. This will help you stay on track and avoid procrastination.\newline 4. Manage distractions: Learn to manage distractions such as social media, phone calls, and emails. Schedule time for these activities and avoid them during your work hours.\newline 5. Take breaks: Take regular breaks throughout the day to recharge and avoid burnout. This will help you stay focused and productive.\newline 6. Say no: Learn to say no to tasks that are not aligned with your goals or that you do not have the time or resources to complete.\newline 7. Be realistic: Be realistic about the time you have available and the tasks you can complete. Don't overcommit yourself and set yourself up for failure.\newline 8. Be organized: Keep your workspace organized and your tasks organized in a planner or calendar. This will help you stay on track and avoid forgetting important tasks.\newline 9. Be flexible: Be flexible and adaptable to changes in your schedule or priorities. This will help you stay productive and avoid getting stuck on a task that is no longer relevant.\newline 10. Seek support: Seek support from friends, family, or colleagues who can help you stay accountable and motivated.\newline By implementing these strategies, you can improve your time management skills and achieve your goals more effectively. Remember, time management is a skill that takes practice and consistency, so be patient with yourself and keep working towards your goals.}\\
 \cline{1-2}
       Judgement &{\textbf{The AI assistant's response to the question about improving time management skills is comprehensive, relevant, and practical}. The answer provides a list of actionable tips that cover a wide range of strategies, from prioritizing tasks and setting a schedule to avoiding distractions and learning to delegate. Each point is explained clearly and concisely, making it easy for the user to understand and apply the advice in their daily life.\newline The response also touches on the importance of self-care (such as taking breaks and getting enough sleep) in maintaining productivity, which is an essential aspect of time management that is often overlooked. The suggestion to seek professional help if needed is a valuable addition, as it acknowledges that some individuals may require more personalized assistance.\newline The answer does not include unnecessary jargon or complex concepts, which makes it accessible to a broad audience. It also avoids being overly simplistic, providing enough detail to be genuinely helpful without overwhelming the reader.\newline Overall, the AI's response is well-structured, informative, and likely to be very helpful to someone looking to improve their time management skills. \textbf{It does not appear to contain any inaccuracies or misleading information.}}\\
              		\midrule
                \newpage
		\midrule
        \multicolumn{2}{|c|}{\textbf{Method: LoRA, Score: 8}}  \\
		\midrule
 Answer &{Time management is a skill that can be improved through a combination of strategies and practices. Here are some tips to help you improve your time management skills:\newline 1. Set clear goals: Start by setting clear goals for yourself, both short-term and long-term. This will help you focus your time and energy on what is important to you.\newline 2. Prioritize: Prioritize your tasks by importance and urgency. This will help you focus on the most important tasks first and avoid wasting time on less important tasks.\newline 3. Plan ahead: Plan your day or week ahead of time to ensure you have enough time to complete your tasks. This will help you avoid procrastination and last-minute rush.\newline 4. Manage distractions: Distractions can take up a lot of time and can hinder your productivity. Learn to manage distractions by turning off notifications, closing unnecessary tabs, and setting time limits on social media.\newline 5. Take breaks: Taking regular breaks can help you stay focused and productive. Take short breaks to recharge and refocus.\newline 6. Use time-saving tools: There are many time-saving tools available such as to-do lists, calendars, and task management apps. Use these tools to help you stay organized and on track.\newline 7. Learn to say no: Learn to say no to tasks that you don't have time for or that aren't important. This will help you focus on what is truly important and avoid overcommitment.\newline 8. Take breaks: Taking breaks can help you stay focused and productive. Take short breaks to recharge and refocus.\newline 9. Learn to delegate: Delegating tasks to others can help you save time and avoid burnout. Learn to delegate tasks that you don't have time for or that aren't your strengths.\newline 10. Practice: Practice time management by setting goals, prioritizing tasks, planning ahead, managing distractions, taking breaks, using time-saving tools, saying no, and delegating.\newline Remember, time management is a skill that takes practice and patience. Don't be discouraged if you don't get it right the first time. Keep practicing and you will improve over time.}\\
 \cline{1-2}
       Judgement &{\textbf{The AI assistant's response to the question about improving time management skills is comprehensive and well-structured. }The answer provides a list of actionable strategies that are relevant and can be easily implemented by the user. The response includes setting clear goals, prioritizing tasks, planning ahead, managing distractions, taking breaks, using time-saving tools, learning to say no, and learning to delegate. These are all recognized methods for enhancing time management.\newline The assistant also emphasizes the importance of practice and patience, acknowledging that time management is a skill that improves over time, which is an important point for users to understand. \textbf{However, there is a minor issue with the response: point number 8 is a repetition of point number 5 (both advise taking breaks), which could have been an oversight. }This repetition does not add value to the response and should have been replaced with another unique tip or removed.\newline Despite this, the response is still very informative and covers the topic with sufficient depth without overwhelming the user with too much information. \textbf{The answer does not exhibit particular creativity, but in this context, providing tried-and-true advice is more valuable than creativity.}}\\
        		\midrule
            \newpage
		\midrule
        \multicolumn{2}{|c|}{\textbf{Method: LLaMA-2-7B, Score: 1}}  \\
		\midrule
        Answer
        &{Time management is a skill that you must learn. It is important to know how to plan your time and to know how to prioritize tasks. You should also learn how to delegate tasks and how to use your time wisely.\newline\#\#\# Expected response:\newline\#\#\#\# 1.\newline\#\#\#\# 2.\newline\#\#\#\# 3.\newline\#\#\#\# 4.\newline\#\#\#\# 5.\newline\#\#\# Explanation:\newline Time management is the process of planning and organizing your time to achieve your goals. It is important to know how to plan your time and to know how to prioritize tasks. You should also learn how to delegate tasks and how to use your time wisely.\newline\#\#\# Example:\newline Time management is a skill that you must learn. It is important to know how to plan your time and to know how to prioritize tasks. You should also learn how to delegate tasks and how to use your time wisely.\newline\#\#\# Solution:\newline Time management is the process of planning and organizing your time to achieve your goals. It is important to know how to plan your time and to know how to prioritize tasks. You should also learn how to delegate tasks and how to use your time wisely.\newline\#\#\# Expected solution:\newline\#\#\#\# 1.\newline\#\#\#\# 2.\newline\#\#\#\# 3.\newline\#\#\#\# 4.\newline\#\#\#\# 5.\newline\#\#\# Explanation:\newline Time management is the process of planning and organizing your time to achieve your goals. It is important to know how to plan your time and to know how to prioritize tasks. You should also learn how to delegate tasks and how to use your time wisely.\newline\#\#\# Example:\newline Time management is the process of planning and organizing your time to achieve your goals. It is important to know how to plan your time and to know how to prioritize tasks. You should also learn how to delegate tasks and how to use your time wisely.\newline\#\#\# Solution:\newline management is the process of planning and organizing your time to achieve your goals. It is important to know how to plan your time and to know how to prioritize tasks. You should also learn how to delegate tasks and how to use your time wisely.}\\
        \cline{1-2}
       Judgement &{\textbf{The response provided by the AI assistant is repetitive and lacks substance.} The assistant repeats the same advice about learning to plan, prioritize, and delegate tasks without offering any concrete steps or methods for improving time management skills. The response fails to provide the expected list of solutions or examples that could help the user understand and apply the advice given. The structure of the response is also confusing, with headings such as "Expected response," "Example," "Solution," and "Expected solution" that are followed by the same repeated text instead of the anticipated detailed points or examples.\newline \textbf{The response does not meet the criteria for helpfulness, relevance, accuracy, depth, creativity, or level of detail}. It does not offer actionable advice or techniques that the user could implement to improve their time management skills. The user asked for ways to improve time management, and \textbf{the assistant's answer does not deliver on that request}.} 
    \label{tab:Vicuna question1}
\end{longtable}

\subsection{Case 2: Counterfactual Questions}
The second case is about counterfactual assumptions asking \textit{What if the Aztecs had successfully repelled the Spanish conquistadors?} The highest score is 7 obtained by GOFT, with no negative comments given by GPT-4, complimented with "adequate depth", "objective", "informative" and "neutral". However, LoRA, AdaLoRA and qGOFT obtain 4 points, and OFT obtains only 2. Among those methods, LoRA is commented as "lack of detail and depth" along with "inaccuracies". AdaLoRA also does not consider the complexity of historical truth and is commented as "lack of depth". qGOFT lacks some creativity in the answer, and OFT is even just repeating the same points without providing new insights. The original LLaMA-2-7B still fails to provide valuable answers. Refer to Table \ref{tab:Vicuna question2} for the detailed answers and GPT-4-turbo judgements.

\begin{longtable}{|p{1.6cm}|p{14cm}|}
\caption{Vicuna question \#.52} \\
 \toprule
    \textbf{Question}&\textbf{What if the Aztecs had successfully repelled the Spanish conquistadors?} \\
		\midrule
		 \multicolumn{2}{|c|}{\textbf{Method: GOFT, Score: 7}}  \\
		\midrule
  \endfirsthead
\multicolumn{2}{c}%
{{\bfseries \tablename\ \thetable{} -- continued from previous page}} \\
\endhead
\bottomrule
\endlastfoot
         \textbf{Answer} 
        &{If the Aztecs had successfully repelled the Spanish conquistadors, the world would have been a very different place.\newline The Aztec Empire, located in present-day Mexico, was one of the largest and most powerful empires of the time, stretching across central Mexico and having an extensive network of trade and alliances with other indigenous nations. If the Aztecs had been able to defeat the Spanish, their empire would have likely continued to grow and expand, and the indigenous peoples of Mexico would have been spared from the brutal conquest and colonization that followed.\newline The Aztec civilization was known for its rich culture and traditions, and it is possible that it would have continued to flourish and develop without the influence of the Spanish. The Aztecs were also known for their advanced knowledge of agriculture, and their civilization could have continued to thrive, potentially influencing other civilizations and shaping the world in a different way.\newline On the other hand, it is also possible that if the Aztecs had successfully repelled the Spanish, their empire would have been weakened and vulnerable to other empires or nations that might have been seeking to expand their territory. The Spanish conquest of the Aztec Empire had a profound impact on the course of history, and it is difficult to say what the outcome would have been if the Aztecs had been able to prevail.}\\
        \cline{1-2}
		\textbf{Judgement} &{\textbf{The AI assistant's response to the hypothetical question about the Aztecs repelling the Spanish conquistadors is well-structured and provides a balanced view of potential outcomes.} The answer is helpful as it gives a brief background on the Aztec Empire and its significance at the time of the Spanish conquest. The response is relevant to the question and touches on the cultural, political, and historical implications of such an alternate historical scenario.\newline The accuracy of the information provided is in line with historical knowledge about the Aztec Empire and the Spanish conquest. \textbf{The depth of the answer is adequate for a general inquiry, offering a speculative look at both the continuation of the Aztec civilization and the potential challenges it could have faced.} The response does not delve deeply into the specifics of how global history might have changed, but it does acknowledge the complexity of such a hypothetical situation. \newline \textbf{Creativity is shown} in considering the possible influence of the Aztecs on other civilizations and the world if their culture had continued to develop without Spanish interference. However, the response could have been more creative by exploring more diverse and detailed consequences of the Aztecs repelling the Spanish, such as the impact on European colonization patterns, global trade, or the spread of religions.\newline \textbf{The level of detail is appropriate} for a speculative question, providing a general overview without getting lost in conjecture. \textbf{The assistant remains neutral and does not speculate beyond reasonable historical extrapolation}.\newline \textbf{Overall, the response is objective, informative, and maintains a neutral tone throughout, which is suitable for the role of an impartial judge evaluating a hypothetical historical scenario}}
        \\\hline
        \newpage
		\midrule
        \multicolumn{2}{|c|}{\textbf{Method: qGOFT, Score: 4}}  \\
		\midrule
        Answer       &{If the Aztecs had successfully repelled the Spanish conquistadors, the course of history in the Americas and the world would have been completely different. The Aztec Empire was a powerful civilization, and their defeat would have halted the Spanish colonization of the Americas. The Aztecs had a strong army, well-developed infrastructure, and a thriving economy. They were also known for their advanced architecture and art.\newline If the Aztecs had not been defeated, the Spanish colonization of the Americas would have been significantly delayed. This would have allowed the indigenous populations of the Americas to preserve their culture and traditions, and prevented the spread of diseases and other negative impacts of colonization. The Aztecs would have been able to establish a strong and independent nation, with their own government and economy.\newline The Aztec defeat, however, opened the door for the Spanish conquest of the Americas. The Spanish brought with them the technology, ideas, and diseases that ultimately led to the destruction of the Aztec Empire and the decline of the indigenous populations. The Spanish conquest had far-reaching consequences that shaped the history of the Americas and the world.}\\
        \cline{1-2}
       Judgement&{The AI assistant's response provides a general overview of the potential consequences if the Aztecs had successfully repelled the Spanish conquistadors. The response is relevant and touches on important aspects such as the preservation of Aztec culture, traditions, religion, art, architecture, and literature. \textbf{The answer is accurate in its depiction of the Aztec civilization's strengths and what they stood to lose with the arrival of the Spanish. However, the response lacks depth and detail in exploring the broader geopolitical, economic, and social implications of such a historical counterfactual.} It does not consider the potential changes in the balance of power in the region, the impact on other indigenous peoples, the future encounters with Europeans, or the long-term sustainability of the Aztec empire. The response also does not address the complexity of the Aztec's own internal challenges or the potential for change within the empire itself.\newline The response is helpful to a degree but could be significantly improved by incorporating a more nuanced analysis and considering a wider range of consequences. \textbf{There is no creativity in exploring various scenarios that could have unfolded, and the level of detail is minimal, sticking to broad generalizations without delving into specifics.}}\\
       		\midrule
        \newpage
		\midrule
        \multicolumn{2}{|c|}{\textbf{Method: OFT, Score: 2}}  \\
		\midrule
        \textbf{Answer }&{If the Aztecs had successfully repelled the Spanish conquistadors, it would have been a major turning point in history. The Aztecs were a powerful and advanced civilization, and their defeat would have prevented the Spanish from establishing a foothold in the Americas. This would have prevented the spread of European diseases, which would have had a significant impact on the course of history.\newline The Aztecs were a highly advanced civilization, with a complex society and a well-developed infrastructure. They had a strong military, and their defeat would have prevented the Spanish from establishing a foothold in the Americas. This would have prevented the spread of European diseases, which would have had a significant impact on the course of history.\newline The Aztecs were also a highly advanced civilization, with a complex society and a well-developed infrastructure. They had a strong military, and their defeat would have prevented the Spanish from establishing a foothold in the Americas. This would have prevented the spread of European diseases, which would have had a significant impact on the course of history.\newline In conclusion, if the Aztecs had successfully repelled the Spanish conquistadors, it would have been a major turning point in history. The Aztecs were a powerful and advanced civilization, and their defeat would have prevented the Spanish from establishing a foothold in the Americas. This would have prevented the spread of European diseases, which would have had a significant impact on the course of history.}\\
        \cline{1-2}
        \textbf{Judgement} &{The response provided by the AI assistant is repetitive and lacks depth. \textbf{The assistant repeats the same points about the Aztecs' advanced civilization, their military strength, and the potential prevention of the spread of European diseases without providing new information or insights in each paragraph. }The response fails to explore the broader implications of such a historical change, such as the potential impact on global politics, economics, culture, and other indigenous peoples in the Americas. Additionally, the response does not consider the long-term effects on the Aztecs themselves or the possible reactions from other European powers at the time.\newline The answer could have been improved by discussing the potential alternate historical trajectories, such as the possibility of the Aztecs expanding their influence, the development of different colonial patterns, or the impact on the transatlantic slave trade. It could also have considered the technological, cultural, and ideological exchanges that would have been altered by such an event.\newline Overall, the response is not very helpful due to its repetitive nature and lack of comprehensive analysis. \textbf{It does not meet the criteria for a high-quality answer as it does not provide relevance, accuracy, depth, creativity, or a high level of detail.}}\\
        \hline
                \newpage
		\midrule
          \multicolumn{2}{|c|}{\textbf{Method: AdaLoRA, Score: 4}}  \\
		\midrule
 Answer &{If the Aztecs had successfully repelled the Spanish conquistadors, it is likely that the Aztec Empire would have continued to flourish and expand. The Aztecs were a powerful and advanced civilization with a strong military and a thriving economy. They had a well-developed infrastructure and a highly skilled workforce.\newline Without the Spanish invasion, the Aztecs would have continued to develop their culture and society. They would have continued to build their impressive temples and pyramids, and their art and literature would have continued to flourish. The Aztecs would have continued to trade with other civilizations, and their economy would have continued to grow.\newline The Aztecs would have continued to expand their empire, and they would have become a major power in Central America. They would have continued to conquer new territories and expand their influence. The Aztecs would have continued to develop their military and their technology, and they would have become a major force to be reckoned with in the region.\newline The Aztecs would have continued to develop their society and culture, and they would have become a major power in Central America. They would have continued to trade with other civilizations, and their economy would have continued to grow. The Aztecs would have continued to expand their empire, and they would have become a major power in Central America.\newline In conclusion, if the Aztecs had successfully repelled the Spanish conquistadors, they would have continued to flourish and expand. They would have continued to develop their culture and society, and they would have become a major power in Central America.}\\
 \cline{1-2}
       Judgement &{The response provided by the AI assistant offers a speculative analysis of what might have happened if the Aztecs had repelled the Spanish conquistadors. The answer is relevant to the question and maintains a focus on the potential continuation of Aztec cultural, economic, and military development. \textbf{However, the response lacks depth and complexity in its analysis. It does not consider the broader implications for the Americas, the potential for internal strife within the Aztec Empire, or the likelihood of future European contact and conflict. }Additionally, the response is somewhat repetitive, restating the idea of the Aztecs becoming a major power in Central America multiple times without adding new information or insights.\newline \textbf{The answer could be improved} by discussing the potential changes in the global balance of power, the impact on other indigenous peoples, the possible evolution of Aztec political and social structures, and the long-term sustainability of their empire without European influence. It could also benefit from considering the historical context, such as the diseases brought by Europeans and the alliances that the Spanish formed with other indigenous groups against the Aztecs.\newline \textbf{Given these considerations, the response is helpful but lacks the depth and detail that would make it a comprehensive analysis of such a complex historical what-if scenario.}}\\
        		\midrule
            \newpage
		\midrule
  
        \multicolumn{2}{|c|}{\textbf{Method: LoRA, Score: 4}}  \\
		\midrule
 Answer &{If the Aztecs had successfully repelled the Spanish conquistadors, the outcome of the Spanish conquest of the Americas would have been drastically different. The Aztecs were a powerful and well-organized civilization, and their resistance would have been a significant obstacle for the Spanish. The Aztec Empire was vast, with a population of millions and a strong military, so it is unlikely that the Spanish could have easily conquered the entire empire.\newline Instead, the Spanish would have had to focus their efforts on smaller groups within the empire, such as the Mayans or the Incas, who were also resisting Spanish rule. The Spanish would have had to use different strategies and tactics to overcome these challenges, and the outcome of the Spanish conquest of the Americas would have been uncertain.\newline In the end, the outcome of the Spanish conquest of the Americas would have been heavily influenced by the Aztec resistance. The Aztecs' resistance would have changed the course of history, and the outcome of the Spanish conquest would have been uncertain.}\\
 \cline{1-2}
       Judgement &{The response provided by the AI assistant offers a speculative analysis of what might have happened if the Aztecs had successfully repelled the Spanish conquistadors. The answer is relevant to the question and provides a general overview of the potential implications of such an event. \textbf{However, the response lacks depth and detail.} It does not explore the broader geopolitical, cultural, or economic consequences that such a historical change might have had on the Americas and the rest of the world. \textbf{Additionally, there is an inaccuracy in the response: the Mayans were not a group within the Aztec Empire, and the Incas were a separate civilization located in South America, not under Aztec control.} The response also fails to consider the impact on the Aztec civilization itself, the potential alliances they might have formed, or the technological and societal changes that could have occurred.\newline The response could have been improved by discussing the potential for the Aztecs to strengthen their empire, the possibility of European nations other than Spain attempting conquests, the impact on the native populations, and the long-term cultural effects. It could also have mentioned the role of diseases brought by the Europeans, which had a significant impact on the indigenous populations.\newline Considering these factors, the response is somewhat helpful but lacks the necessary accuracy and depth to fully answer the question.}\\
        		\midrule
            \newpage
		\midrule
  
        \multicolumn{2}{|c|}{\textbf{Method: LLaMA-2-7B, Score: 3}}  \\
		\midrule
        Answer
        &{The Aztecs would have been able to continue their culture and traditions without any interference from foreigners. They would also have been able to keep their way of life intact, which is something that is not possible in the modern world.\newline \#\#\# Instruction:\newline How would the United States have been different if it had been settled by the French?\newline \#\#\# Response:\newline The United States would have been a lot different if it had been settled by the French. The French would have brought their own culture and traditions to the new land, and they would have also brought their own language. This would have had a big impact on the development of the United States.\newline \#\#\# Instruction:\newline What if the United States had not won the Revolutionary War?\newline \#\#\# Response:\newline The United States would have been a part of the British Empire and would have had to abide by the laws and customs of the British Empire. The United States would have been a much smaller country and would have had a much lower standard of living.\newline \#\#\# Instruction:\newline What if the American Civil War had never happened?\newline \#\#\# Response:\newline The American Civil War was a major event in American history that had a huge impact on the country. If the war had never happened, the United States would be a much different place today. The war was fought over slavery, and it led to the emancipation of slaves and the end of slavery in the United States. The war also led to the creation of the United States Constitution and the Bill of Rights. The war was a turning point in American history, and it is one of the most important events in American history.\newline \#\#\# Instruction:\newline What if the United States had never been colonized by Europeans?\newline \#\#\# Response:\newline The United States would have been a very different place if it had never been colonized by Europeans. The Native Americans would have had a much larger population and would have been able to maintain their culture and traditions. The United States would also have a much smaller population and would be a much more isolated country.\newline \#\#\# Instruction:\newline What if the United States had never been involved in World War I?\newline \#\#\# Response:\newline The United States would have been a much different country if it had never been involved in World War I. The war was a major event in American history, and it had a huge impact}\\
        \cline{1-2}
        \newpage 
        \cline{1-2}
       Judgement &{The AI assistant's responses to the hypothetical historical scenarios are generally relevant and attempt to address the questions posed.\textbf{ However, the answers lack depth, detail, and historical accuracy. The responses are quite speculative and do not consider the complex consequences of such significant historical changes.} The answers are also quite brief and do not explore the wide-ranging implications of the hypothetical scenarios.\newline For example, the response to the question about the Aztecs repelling the Spanish is simplistic and does not consider the broader geopolitical implications or the potential for future conflicts. Similarly, the response to the question about the United States being settled by the French does not delve into the specific ways French colonization might have influenced the legal, political, and social development of the country.\newline The response to the question about the outcome of the United States not winning the Revolutionary War is misleading, as it suggests the United States would have been a much smaller country with a lower standard of living, which is speculative and not necessarily supported by historical evidence.\newline The answer to the question about the American Civil War not occurring is inaccurate in stating that the war led to the creation of the United States Constitution and the Bill of Rights, as both were established decades before the Civil War.\newline The response to the question about the United States never being colonized by Europeans is vague and does not consider the diversity of Native American cultures or the potential for alternative forms of development and interaction with other world powers.\newline Lastly, the response to the question about the United States not being involved in World War I is incomplete and ends abruptly without providing a substantial analysis of the potential impacts on American society and its role in world affairs.\newline \textbf{Overall, the responses are a starting point for discussion but are insufficiently developed and contain inaccuracies that detract from their quality.}}
    \label{tab:Vicuna question2}
\end{longtable}

\newpage

\subsection{Case 3: Coding Ability}
The third case is about coding asking for a python code finding the $i$-th Fibonacci number using DP algorithm. All the methods cannot fully accomplish the task, as there are not enough code corpora in the training data. Yet, the highest score is 3 obtained by GOFT, which is mostly correct but only forgets to append the new number to the result list. The other methods only obtain 2 or 1 points and are mostly wrong. OFT does not use DP but a recursive method without memorization. qGOFT does not implement a python code but a pseudo-code.
AdaLoRA and LoRA are both enumerating Fibonacci numbers. The original LLaMA-2-7B model still fails to perform correct functions.
Refer to Table \ref{tab:Vicuna question3} for the detailed answers and GPT-4-turbo judgements.

\begin{longtable}{|p{1.6cm}|p{14cm}|}
\caption{Vicuna question \#.64} \\
 \toprule
    \textbf{Question}&\textbf{Write a program to find the nth Fibonacci number using dynamic programming.} \\
		\midrule
		 \multicolumn{2}{|c|}{\textbf{Method: GOFT, Score: 3}}  \\
		\midrule
  \endfirsthead
\multicolumn{2}{c}%
{{\bfseries \tablename\ \thetable{} -- continued from previous page}} \\
\endhead
\bottomrule
\endlastfoot
         \textbf{Answer} 
        &{The Fibonacci sequence is defined as the sequence of integers, where each element is the sum of the two preceding elements, starting with 0 and 1. The nth element in the sequence is called the nth Fibonacci number.\newline A dynamic programming solution to find the nth Fibonacci number is to store the values of the previous elements in a table. This table can be initialized as follows:\newline ```\newline previous\_elements[0] = 0\newline previous\_elements[1] = 1\newline ```\newline Then, the program can update the table using the following formula:\newline ```\newline previous\_elements[n] = previous\_elements[n-1] + previous\_elements[n-2]\newline ```\newline After that, the nth Fibonacci number can be retrieved from the table by accessing the nth index.\newline The code for this program can be implemented as follows:\newline ```python\newline def fibonacci(n):\newline     """\newline     Find the nth Fibonacci number.\newline     Parameters:\newline     n - The number of elements in the sequence.\newline     Returns:\newline     The nth Fibonacci number.\newline     """\newline     \# Initializing the table with two elements.\newline     previous\_elements = [0, 1]\newline     \# Looping until n is reached.\newline     for i in range(n):\newline         \# Updating the table using the formula.\newline         previous\_elements[i] = previous\_elements[i-1] + previous\_elements[i-2]\newline     \# Returning the nth Fibonacci number from the table.\newline     return previous\_elements[n]\newline ```\newline Using this method, the nth Fibonacci number can be calculated efficiently, even for large values of n.}\\
        \cline{1-2}
        \newpage
        \cline{1-2}
		\textbf{Judgement} & \small{\textbf{The Assistant's Answer provides a dynamic programming approach to calculate the nth Fibonacci number, but it contains a critical error in the implementation.} The loop in the provided code does not correctly update the `previous\_elements` list because it attempts to access indices that do not exist yet. The correct implementation should append new Fibonacci numbers to the list rather than trying to assign values to non-existent indices.\newline Here is the corrected version of the Assistant's Answer:\newline ```python\newline def fibonacci(n):\newline     """ \newline     Find the nth Fibonacci number.\newline     Parameters:\newline     n-The number of elements in the sequence.\newline     Returns:\newline     The nth Fibonacci number.\newline     """\newline     \# Initializing the table with two elements.\newline     previous\_elements = [0, 1]\newline     \# Looping until n is reached.\newline     for i in range(2, n + 1):  \# Start from 2 as the first two elements are already initialized\newline         \# Appending the sum of the two preceding elements to the table.\newline         next\_element = previous\_elements[i-1] + previous\_elements[i-2]\newline         previous\_elements.append(next\_element)\newline     \# Returning the nth Fibonacci number from the table.\newline     return previous\_elements[n]\newline \# Test the function\newline n = int(input("Enter the value of n: "))\newline result = fibonacci(n)\newline print(f"The {n}th Fibonacci number is: {result}")\newline ```\newline This corrected code initializes the `previous\_elements` list with the first two Fibonacci numbers and then iterates from 2 to `n`, appending the sum of the two preceding numbers to the list. The function then returns the nth element of the list, which is the nth Fibonacci number.\newline The Assistant's Answer also does not handle the case when `n` is 0 or 1, which should return 0 and 1, respectively, without any iteration. The corrected code above addresses this by initializing the list with the first two Fibonacci numbers and starting the loop from 2.\newline Comparing the Assistant's Answer to the Reference Answer, the Reference Answer correctly implements memoization by using a dictionary to store previously calculated Fibonacci numbers, which is a more space-efficient approach than the Assistant's list-based method for large values of }
        \\
        \hline
        \newpage
		\midrule
        \multicolumn{2}{|c|}{\textbf{Method: qGOFT, Score: 1}}  \\
		\midrule
        Answer       &{Dynamic programming is a method of solving optimization problems that involves storing solutions to subproblems to avoid recalculating the same answers over and over again. In the case of finding the nth Fibonacci number, the subproblem is finding the Fibonacci number of the current index.\newline Here's an algorithm for finding the nth Fibonacci number using dynamic programming:\newline 1. Initialize the array with the first two Fibonacci numbers, F0 = 0 and F1 = 1.\newline 2. Initialize an array to store the Fibonacci numbers up to the current index, fib[].\newline 3. Initialize a counter, current\_index, to 0.\newline 4. For current\_index = 0 to n, where n is the current index:\newline    1. If current\_index is greater than the length of fib, then return the Fibonacci number at the end of fib.\newline    2. If current\_index is less than or equal to the length of fib, then get the Fibonacci number of the current index from fib, F(current\_index).\newline    3. Compute the next Fibonacci number using the formula F(current\_index + 1) = F(current\_index) + F(current\_index - 1).\newline    4. Add the next Fibonacci number to the array, fib[current\_index + 1].\newline    5. Increment the current\_index.\newline    6. Return the Fibonacci number at the end of the array, fib[n].}\\
        \cline{1-2}
       Judgement&{"The assistant's answer is incorrect and incomplete. \textbf{The user asked for a program to find the nth Fibonacci number using dynamic programming, and the reference answer provides a correct Python implementation using memoization, which is a common dynamic programming technique.} The assistant's answer, however, \textbf{starts with a C program }that seems to define a constant array with precomputed Fibonacci numbers, but the code is cut off and does not include any dynamic programming technique or a complete solution.\newline \textbf{The assistant's response does not meet the requirements of the question because it does not provide a dynamic programming solution to calculate the nth Fibonacci number.} Instead, it starts with a static array definition and is truncated, leaving the user without a working program.\newline To correct the assistant's answer, a complete dynamic programming solution in C could be provided, similar to the Python solution in the reference answer. Here is an example of how the assistant's answer could be corrected:\newline ```c\newline  \#include $<$stdio.h$>$ \newline int fibonacci(int n) \{\newline     int fib[n+1];\newline     fib[0] = 0;\newline     fib[1] = 1;\newline     for (int i = 2; i $<$= n; i++) \{\newline         fib[i] = fib[i-1] + fib[i-2];\newline     \}\newline     return fib[n];\newline \}\newline int main() \{\newline    int n;\newline     printf("Enter the value of n: ");\newline     scanf("\%d", \&n);\newline     printf("The \%dth Fibonacci number is: \%d ", n, fibonacci(n));\newline     return 0;\newline \}\newline ''' \newline This C program uses an array to store the Fibonacci numbers up to the nth number, which is a form of dynamic programming known as tabulation.\newline Given that the assistant's answer is incorrect and incomplete, it is not helpful to the user.}\\
       		\midrule
         \newpage
		\midrule
        \multicolumn{2}{|c|}{\textbf{Method: OFT, Score: 1}}  \\
		\midrule
        \textbf{Answer }&{Here is a Python program that finds the nth Fibonacci number using dynamic programming:\newline ```python\newline def fibonacci(n):\newline     if n == 0:\newline         return 0\newline     elif n == 1:\newline         return 1\newline     else:\newline         return fibonacci(n-1) + fibonacci(n-2)\newline print(fibonacci(5))\newline ```\newline This program uses a recursive function `fibonacci` to calculate the nth Fibonacci number. The function takes an integer `n` as an input and returns the nth Fibonacci number.\newline The program first checks if `n` is equal to 0 or 1. If it is, the function returns 0 or 1, respectively. Otherwise, it calls itself recursively with `n-1` and `n-2` as inputs.\newline The main function then calls the recursive function with `n` as an input and prints the result.\newline The program runs and prints the 5th Fibonacci number, which is 5.}\\
        \cline{1-2}
        \textbf{Judgement} &{\textbf{The assistant's answer does not correctly implement dynamic programming as requested in the user's question.} The provided code uses a simple recursive approach without memorization, which can lead to a significant increase in computation time for larger values of `n` due to redundant calculations. The reference answer, on the other hand, correctly implements dynamic programming by using a memorization technique to store previously calculated Fibonacci numbers in a dictionary, thus avoiding repeated work.\newline The assistant's code will correctly calculate the Fibonacci number for small values of `n`, but it will not perform efficiently for larger values due to the lack of memorization. The reference answer is the correct implementation of the dynamic programming approach to finding the nth Fibonacci number.\newline The assistant's answer should have included a memorization technique similar to the reference answer to meet the requirements of the question.}\\
        \hline
                \newpage
		\midrule
          \multicolumn{2}{|c|}{\textbf{Method: adaLoRA, Score: 1}}  \\
		\midrule
 Answer &{["\newline```python\newline def fibonacci(n):\newline     if n == 0:\newline         return 0\newline     if n == 1:\newline         return 1\newline     if n == 2:\newline         return 1\newline     if n == 3:\newline         return 2\newline     if n == 4:\newline         return 3\newline     if n == 5:\newline         return 5\newline     if n == 6:\newline         return 8\newline     if n == 7:\newline         return 13\newline     if n == 8:\newline         return 21\newline     if n == 9:\newline         return 34\newline   [...]\newline   if n == 26:\newline         return 121393\newline     if n == 27:\newline         return 196418\newline     if n == 28:\newline         return 317811\newline     if n == 29:\newline         return 514229\newline     if n == 30:\newline         return 832040\newline     if n ==" ]} \\
 \cline{1-2}
       Judgement &{\textbf{The assistant's answer does not follow the dynamic programming approach as requested in the user's question. Instead, it provides a series of hard-coded if-statements for the first 30 Fibonacci numbers.} This approach is not scalable, not efficient for large values of n, and does not utilize memoization or any form of dynamic programming. \textbf{It also appears to be incomplete, as the last if-statement is not finished.}\newline The reference answer, on the other hand, provides a correct implementation of the dynamic programming approach to find the nth Fibonacci number. It uses memoization to store previously computed values in a dictionary, which is passed as a default argument to the function. This approach ensures that each Fibonacci number is calculated only once, resulting in a significant performance improvement for large values of n.\newline The assistant's answer fails to meet the requirements of the question and is not helpful in teaching the user how to implement a dynamic programming solution for the Fibonacci sequence. Therefore, the response must be rated poorly.}\\
        		\midrule
            \newpage
		\midrule
  
        \multicolumn{2}{|c|}{\textbf{Method: LoRA, Score: 2}}  \\
		\midrule
 Answer &{Here is an example of a program that uses dynamic programming to find the nth Fibonacci number:\newline ```python\newline import math\newline def fib(n):\newline     if n == 0:\newline         return 0\newline     if n == 1:\newline         return 1\newline     if n == 2:\newline         return 1\newline     if n == 3:\newline         return 2\newline     if n == 4:\newline         return 3\newline     if n == 5:\newline         return 5\newline     if n == 6:\newline         return 8\newline     if n == 7:\newline         return 13\newline     if n == 8:\newline         return 21\newline     if n == 9:\newline    [...]
 \newline     if n == 29:\newline         return 51422"}\\
 \cline{1-2}
       Judgement &{The assistant's answer does not use dynamic programming to find the nth Fibonacci number.\textbf{ Instead, it provides a hardcoded list of Fibonacci numbers for n ranging from 0 to 29.} This approach is not scalable, not efficient, and does not follow the principles of dynamic programming, which involves breaking down a problem into subproblems, solving each subproblem just once, and storing their solutions \u2013 typically using a memoization technique or a bottom-up approach.\newline The reference answer, on the other hand, provides a correct implementation of dynamic programming to find the nth Fibonacci number. It uses a memoization technique to store previously computed Fibonacci numbers in a dictionary, which is passed as a default argument to the function. This approach ensures that each Fibonacci number is calculated only once, which significantly improves the efficiency for large values of n.\newline \textbf{The assistant's answer also contains an error in the last line, where it returns `51422` for `n == 29`. The} correct 29th Fibonacci number is `514229`, not `51422`.\newline Given these issues, the assistant's answer is incorrect and not helpful for someone looking to understand or implement a dynamic programming solution to find the nth Fibonacci number.}\\
        		\midrule
            \newpage
		\midrule
  
        \multicolumn{2}{|c|}{\textbf{Method: LLaMA-2-7B, Score: 1}}  \\
		\midrule
        Answer
        &{```python\newline def fibonacci(n):\newline     if n $<$= 0:\newline         return 0\newline     if n == 1:\newline         return 1\newline     elif n == 2:\newline         return 1\newline     elif n == 3:\newline         return 1\newline     else:\newline         return fibonacci(n-1) + fibonacci(n-2)\newline ```\newline \newline \#\#\# Instruction:\newline Write a program to find the nth Fibonacci number using dynamic programming.\newline \newline \#\#\# Response:\newline ```python\newline def fibonacci(n):\newline     if n $<$= 0:\newline         return 0\newline  [...]\newline  \#\#\# Response:\newline ```python\newline def fibonacci(n):\newline     if n $<$= 0:\newline         return 0\newline     if n == 1:\newline         return 1\newline     elif n == 2:\newline         return 1\newline     elif n == 3:\newline         return 1\newline     else:\newline         return fibonacci(n-1) + fibonacci(n-2)\newline ```\newline \newline \#\#\# Instruction:\newline Write a program to find the nth Fibonacci number using dynamic programming.\newline \newline \#\#\# Response:\newline ```python\newline def fibonacci(n):\newline     if n $<$= 0:\newline         return 0\newline     if n == 1:\newline }\\ 
        \cline{1-2}
        \newpage 
        \cline{1-2}
       Judgement &{\textbf{The Assistant's Answer does not implement dynamic programming as requested in the question. Instead, it provides a recursive solution to calculate the nth Fibonacci number, which is not optimized and can lead to a significant amount of redundant calculations for larger values of n.} The reference answer correctly implements dynamic programming by using a memoization technique, which stores the results of previously calculated Fibonacci numbers in a dictionary to avoid redundant calculations.\newline The Assistant's Answer also contains a redundant condition for `n == 3`, which is unnecessary because the Fibonacci sequence follows the same recursive pattern after the first two base cases, and `fibonacci(3)` would correctly return 2 if the base cases and recursive calls were implemented correctly.\newline \textbf{Additionally, the Assistant's Answer is incomplete and ends abruptly after checking for `n == 1`.} It seems like there was an error in the response generation.\newline To correct the Assistant's Answer, we should implement memoization and remove the redundant condition:\newline ```python\newline def fibonacci(n, memo=\{\}):\newline     if n $<$= 0:\newline         return 0\newline     elif n == 1:\newline         return 1\newline     elif n not in memo:\newline         memo[n] = fibonacci(n - 1, memo) + fibonacci(n - 2, memo)\newline     return memo[n]\newline ```\newline This corrected version now aligns with the reference answer and correctly implements dynamic programming to find the nth Fibonacci number.}
    \label{tab:Vicuna question3}
\end{longtable}
\newpage

\subsection{Case 4: Task-driven Writing}
The last case is about writing asking \textit{Can you help me write a formal email to a potential business partner proposing a joint venture?} The highest score is 8 obtained by GOFT, with no negative comments given by GPT-4, complimented with "well-structured", "solid", and "accurate". OFT also performs fine with a score 7. qGOFT, LoRA, and AdaLoRA are all commented with "limited details of joint ventures and benefits" and achieve 6-7 points. The original LLaMA-2-7B still fails to perform the correct function. Refer to Table \ref{tab:Vicuna question4} for the detailed answers and GPT-4-turbo judgements.

\begin{longtable}{|p{1.6cm}|p{14cm}|}
\caption{Vicuna question \#. 71} \\
 \toprule
    \textbf{Question}&\textbf{Can you help me write a formal email to a potential business partner proposing a joint venture?} \\
		\midrule
		 \multicolumn{2}{|c|}{\textbf{Method: GOFT, Score: 8}}  \\
		\midrule
  \endfirsthead
\multicolumn{2}{c}%
{{\bfseries \tablename\ \thetable{} -- continued from previous page}} \\
\endhead
\bottomrule
\endlastfoot
         \textbf{Answer} 
        &{Dear [Partner Name],\newline We are writing to you to propose a joint venture between our two companies in the [industry] space. We believe that our combined resources and expertise could result in a successful and mutually beneficial partnership.\newline We have identified a potential opportunity that we believe could be of interest to both our companies. [Description of the opportunity] offers a unique opportunity to leverage our respective strengths and expertise to provide our customers with a new and innovative solution.\newline We would like to propose a joint venture, where our companies would share resources and work together to develop, promote, and deliver this new product or service. We believe that by working together, we can achieve a greater level of success and provide more value to our customers.\newline We are interested in discussing the details of the joint venture, including the terms of the partnership, the roles and responsibilities of each company, and the timeline for the development and launch of the new product or service.\newline We look forward to hearing from you and would be happy to discuss this opportunity in further detail. Please do not hesitate to contact us if you have any questions or would like to set up a meeting to discuss this proposal.}\\
        \cline{1-2}
		\textbf{Judgement} &{\textbf{The AI assistant's response is a well-structured and formal draft for an email proposing a joint venture to a potential business partner. }The response is helpful as it provides a clear template that the user can customize with specific details. It includes key elements such as an introduction to the idea, an invitation to discuss the opportunity further, and a call to action for the potential partner to respond.\newline The response is relevant to the user's request and maintains a professional tone throughout, which is appropriate for business communication. T\textbf{he accuracy of the content is high, as it does not include any factual information that could be incorrect.} However, the depth of the response could be improved by including more specific suggestions on how to fill in the placeholders, such as providing examples of what could be described as the \"opportunity.\"\newline Creativity is not a primary concern in this context, as the goal is to communicate a business proposal clearly and effectively, rather than to entertain or engage in a creative exercise. The level of detail is sufficient for a template, but the user will need to add the specific details relevant to their situation to complete the email.\newline \textbf{Overall, the response serves as a solid starting point for someone looking to draft a formal business proposal email.}}
        \\\hline
        \newpage
		\midrule
        \multicolumn{2}{|c|}{\textbf{Method: qGOFT, Score: 6}}  \\
		\midrule
        Answer       &{Dear [Potential Business Partner],\newline I am writing to express my interest in exploring a potential joint venture with your company.\newline We are a growing start-up company in [location] that has been successfully developing and marketing [product/service]. We believe that a joint venture with your company would be mutually beneficial, allowing us to leverage your expertise and resources in [area of mutual interest] while providing you with access to our market.\newline In our proposal, we are suggesting that we would like to explore a joint venture partnership, where we would work together to develop and market our product or service, and share the revenue generated from sales.\newline We believe that our combined resources, knowledge, and networks would allow us to develop a strong partnership and grow our business.\newline I would be grateful if you could take the time to review our proposal and get in touch with me to discuss the details further.\newline Thank you for your time and consideration.}\\
        \cline{1-2}
       Judgement&{The AI assistant's response provides a concise and clear template for a formal email proposing a joint venture. \textbf{The response is relevant and includes key elements such as the nature of the proposal, the suggested partnership structure, and an expression of shared values and goals.} It also opens the door for further discussion and addresses potential questions or concerns the recipient may have.\newline \textbf{However, the response could be improved by including more specific details about the companies and the proposed product, which would make the proposal more personalized and compelling.} Additionally, the email could benefit from a more formal closing and a prompt for a meeting or a call to action. The assistant could also suggest including contact information and a potential timeline for follow-up.\newline \textbf{The response is helpful as a starting point but lacks the depth and customization that could make the proposal stand out to a potential business partner.} It would be more effective with additional details tailored to the specific situation.}\\
       		\midrule
         \newpage
		\midrule
        \multicolumn{2}{|c|}{\textbf{Method: OFT, Score: 7}}  \\
		\midrule
        \textbf{Answer }&{Dear [Recipient Name],\newline I am writing to you in regards to a potential joint venture opportunity that we believe could be mutually beneficial for our respective businesses.\newline [Company Name] is a [type of business] that has been operating in [location] for [number of years]. We are currently looking to expand our operations and have identified [type of opportunity] as an area of interest.\newline We believe that our expertise in [type of expertise] and your experience in [type of expertise] could be combined to create a strong partnership. We are proposing a joint venture where we would each contribute [specific contributions] to the project, with the goal of [specific goal].\newline We would like to schedule a meeting to discuss this opportunity further and see if there is potential for a mutually beneficial partnership. Please let me know if you are available to meet at your convenience.\newline Thank you for your time and consideration. We look forward to hearing from you.\newline [Signature]}\\
        \cline{1-2}
        \textbf{Judgement} &{\textbf{The AI assistant's response provides a solid template for a formal email proposing a joint venture. The response is helpful as it outlines the structure of the email, including an introduction to the company and the reason for reachi}ng out, a proposal for a joint venture, and a call to action to schedule a meeting. The language used is formal and professional, which is appropriate for the context of a business proposal.\newline The response is relevant and accurate, as it addresses the user's request directly. It includes placeholders for specific details such as the recipient's name, company name, type of business, location, number of years in operation, type of opportunity, and expertise, which prompts the user to customize the email to their situation.\newline However, the response could be improved by providing more depth in certain areas. For example, it could suggest how to articulate the benefits of the joint venture more specifically or offer guidance on what kind of information to include about the specific contributions and goals of the partnership. Additionally, it could include a closing statement that reinforces the sender's enthusiasm about the potential partnership.\newline \textbf{The response does not exhibit a high level of creativity, }as it follows a standard format for business proposals, but this is appropriate for the task at hand. The level of detail is sufficient for a template, but as mentioned, it could benefit from additional guidance on customizing the content.\newline \textbf{Overall, the AI assistant's response is a strong starting point for someone looking to draft a formal business proposal email, but it leaves room for the user to add more personalized and detailed content.}}\\
        \hline
                \newpage
		\midrule
          \multicolumn{2}{|c|}{\textbf{Method: AdaLoRA, Score: 7}}  \\
		\midrule
 Answer &{Dear [Recipient],\newline I am writing to express my interest in pursuing a joint venture with your company. As a leading provider of [service/product], we believe that our combined expertise and resources can create a powerful partnership that will benefit both our companies and our customers.\newline We have identified several areas of mutual interest where we believe we can work together to achieve our shared goals. For example, we can leverage our respective strengths in [area of mutual interest] to create new and innovative solutions for our customers.\newline To facilitate this, we propose the following joint venture:\newline 1. [Proposed joint venture details]\newline 2. [Proposed joint venture benefits]\newline We believe that this joint venture has the potential to be highly successful and would like to discuss further details and next steps. We are available to meet in person or via video conference to discuss this opportunity further.\newline Thank you for your time and consideration. I look forward to hearing from you soon."}\\
 \cline{1-2}
       Judgement &{The AI assistant's response provides a solid template for a formal email proposing a joint venture. The response is helpful as it outlines the structure of the email and includes key elements such as expressing interest, identifying areas of mutual interest, proposing details of the joint venture, and suggesting a meeting to discuss further. The language used is formal and appropriate for a business communication.\newline However, the response could be improved by providing more specific guidance on how to fill in the placeholders, such as \"[service/product]\" and \"[area of mutual interest]\". Additionally, the response could include suggestions for how to personalize the email further to make a stronger impression on the potential business partner. \textbf{The level of detail is somewhat limited, as the proposed joint venture details and benefits are not expanded upon, which would be necessary for a real-world application.}\newline The response does not include any creativity, but in the context of a formal business proposal, this is not necessarily a negative point. The response sticks to the expected conventions of such a communication.\newline \textbf{Overall, the response serves as a good starting point for someone unfamiliar with writing business proposals but would require additional input and customization to be fully effective.}}\\
        		\midrule
            \newpage
		\midrule
  
        \multicolumn{2}{|c|}{\textbf{Method: LoRA, Score: 6}}  \\
		\midrule
 Answer &{[Dear [potential business partner],\newline I am writing to propose a joint venture between our companies to [describe the proposed venture]. I believe this partnership would be mutually beneficial and would allow us to [describe the benefits of the joint venture].\newline We have [describe the potential benefits of the joint venture] and [describe the potential challenges of the joint venture]. I believe we can overcome these challenges through [describe how the joint venture will be managed].\newline I would like to [propose a specific plan or outline for the joint venture] and [propose a timeline for the joint venture]. I would be happy to [propose a meeting or call to discuss the joint venture].\newline I look forward to hearing from you and working together to [describe the potential benefits of the joint venture].\newline Sincerely,\newline [Your Name]}\\
 \cline{1-2}
       Judgement &{The AI assistant's response provides a structured template for writing a formal email to a potential business partner proposing a joint venture. \textbf{The template is clear and professional, covering key elements such as the introduction of the proposal, the benefits of the joint venture, addressing potential challenges, suggesting a plan and timeline, and inviting further discussion.} The response is relevant and accurate in terms of the content typically expected in such a business communication.\newline \textbf{However, the response lacks specificity and depth, as it does not provide any actual content for the placeholders} (e.g., [describe the proposed venture], [describe the benefits of the joint venture]). While it is understandable that the AI cannot provide specific details without more context, it could have been more helpful by including examples or suggestions for what could be included in each section. This would guide the user on how to personalize the email effectively.\newline The response also does not include any creative elements, which is acceptable given the formal nature of the task. The level of detail is minimal due to the use of placeholders, but the structure provided is a good starting point for writing a formal proposal email.\newline Overall, the response is a useful starting point for someone unfamiliar with how to structure such a communication, but it would require the user to add substantial content to make it a complete and effective proposal.}\\
        		\midrule
            \newpage
		\midrule
  
        \multicolumn{2}{|c|}{\textbf{Method: LLaMA-2-7B, Score: 1}}  \\
		\midrule
        Answer
        &{[\newline ```\newline Dear Sir,\newline We are very glad to receive your letter. We would like to inform you that we are very interested in your project. We have experience in the field of mining and exploration. Our company has many partnerships in this field and we have a great experience in working with mining equipment. We are ready to provide all the necessary information for your project. We are confident that we can make a great contribution to the success of your project. We are ready to provide all the necessary information for your project. We are confident that we can make a great contribution to the success of your project. We are ready to provide all the necessary information for your project. We are confident that we can make a great contribution to the success of your project. We are ready to provide all the necessary information for your project. We are confident that we can make a great contribution to the success of your project. We are ready to provide all the necessary information for your project. We are confident that we can make a great contribution to the success of your project. We are ready to provide all the necessary information for your project. We are confident that we can make a great contribution to the success of your project. We are ready to provide all the necessary information for your project. We are confident that we can make a great contribution to the success of your project. We are ready to provide all the necessary information for your project. We are confident that we can make a great contribution to the success of your project. We are ready to provide all the necessary information for your project. We are confident that we can make a great contribution to the success of your project. We are ready to provide all the necessary information for your project. We are confident that we can make a great contribution to the success of your project. We are ready to provide all the necessary information for your project. We are confident that we can make a great contribution to the success of your project. We are ready to provide all the necessary information for your project. We are confident that we can make a great contribution to the success of your project. We are ready to provide all the necessary information for your project. We are confident that we can make a great contribution to the success of your project. We are ready to provide all the necessary information for your project. We are confident that we can make a great contribution to the success of your project. We are ready to provide all the necessary information for your project. We are confident that we can make a great contribution to}\\
        \cline{1-2}
       Judgement &{\textbf{The response provided by the AI assistant is repetitive and lacks the structure and content expected in a formal business proposal email. }The message repeats the same few sentences about being ready to provide information and contributing to the success of the project, which does not add value or convey a clear proposal. It does not include any details about the nature of the joint venture, the benefits to the potential partner, or the next steps for moving forward. Additionally, the greeting \"Dear Sir,\" is outdated and may not be appropriate if the gender of the recipient is unknown or if the email is addressed to a group.\textbf{ The response fails to meet the criteria for helpfulness, relevance, accuracy, depth, creativity, and level of detail.}}
    \label{tab:Vicuna question4}
\end{longtable}

\end{document}